\documentclass[10pt,twocolumn,letterpaper]{article}

\usepackage[pagenumbers]{cvpr} % To force page numbers, e.g. for an arXiv version

\definecolor{cvprblue}{rgb}{0.21,0.49,0.74}
\usepackage[pagebackref,breaklinks,colorlinks,allcolors=cvprblue]{hyperref}
\usepackage{comment}
\usepackage{amsmath}
\usepackage{amssymb}
\usepackage{booktabs} % For \toprule, \midrule, \cmidrule, \bottomrule
\usepackage{graphicx} % For \resizebox
\usepackage[table]{xcolor} % For \rowcolor
\usepackage{xcolor}
\usepackage{tikz}
\usepackage{pgfplots}
\usepackage{url}
\usepackage{multirow}
\usepgfplotslibrary{groupplots}
\pgfplotsset{compat=1.18}
\definecolor{colorOurs}{RGB}{217, 30, 24}      % Highlight Red/Coral (Selected / Ours)
\definecolor{colorNaive}{RGB}{127, 140, 141}   % Neutral Gray (Naive Baseline)
\definecolor{colorBlur}{RGB}{41, 128, 185}     % Blue (2D Blur)
\definecolor{colorRadio}{RGB}{39, 174, 96}     % Green (Radiometric)
\definecolor{colorStride10}{RGB}{142, 68, 173}  % Purple (Stride 10)
\definecolor{colorSpatial}{RGB}{230, 126, 34}   % Orange (Spatial Noise)

\def\paperID{497} % *** Enter the Paper ID here
\def\confName{3DV\xspace}
\def\confYear{2027\xspace}

\title{Rethinking 3D Noise: Learning 3D-Aware Video Priors via Optimization-Free Morphological Perturbations}

\author{
  Onat Şahin\textsuperscript{1,2} \quad 
  Mohammad Altillawi\textsuperscript{2} \quad 
  George Eskandar\textsuperscript{3} \quad 
  Carlos Carbone\textsuperscript{2} \quad 
  Ziyuan Liu\textsuperscript{2}  \\ \\
  \textsuperscript{1}Technical University of Munich \quad 
  \textsuperscript{2}Huawei Heisenberg Research Center \quad
  \textsuperscript{3}Tavus \\
}

\begin{document}
\maketitle
\begin{abstract}
3D scene representations like NeRF and 3D Gaussian Splatting (3DGS) suffer severe artifacts in sparse-view settings. Recent generative 3D artifact fixers attempt to address this, but rely on paired corrupted and clean renders requiring costly, per-scene reconstructions across varying view configurations. While 2D image augmentations act as instant regularizers, no explicit equivalents exist for 3D representations to preserve spatial consistency across views, an essential property for 3D-aware training. We propose \textbf{3D Morphological Perturbations} as an optimization-free regularizer that preserves spatial consistency. Leveraging explicit 3DGS, we treat each Gaussian as a fundamental building block—analogous to a 2D pixel—and apply perturbations across its morphological parameter space via scale, rotation, and pruning. Our method eliminates per-scene 3DGS optimization loops from dataset curation while enabling models to learn stronger geometric priors than sparse-view baselines in diagnostic ablations conducted on a lightweight video diffusion sandbox. Scaled to a 14B-parameter video model via ControlNet, our approach maintains visual fidelity while reducing mean depth error by 12.5\% over state-of-the-art image-to-image 3D artifact refiners, ultimately boosting downstream robotics policy success rates by up to 8.0\% across 3 of 4 manipulation tasks.

\end{abstract}    
\vspace{-10pt}
\section{Introduction}
\label{sec:intro}

\begin{figure}[t]
  \centering
  \includegraphics[width=\columnwidth]{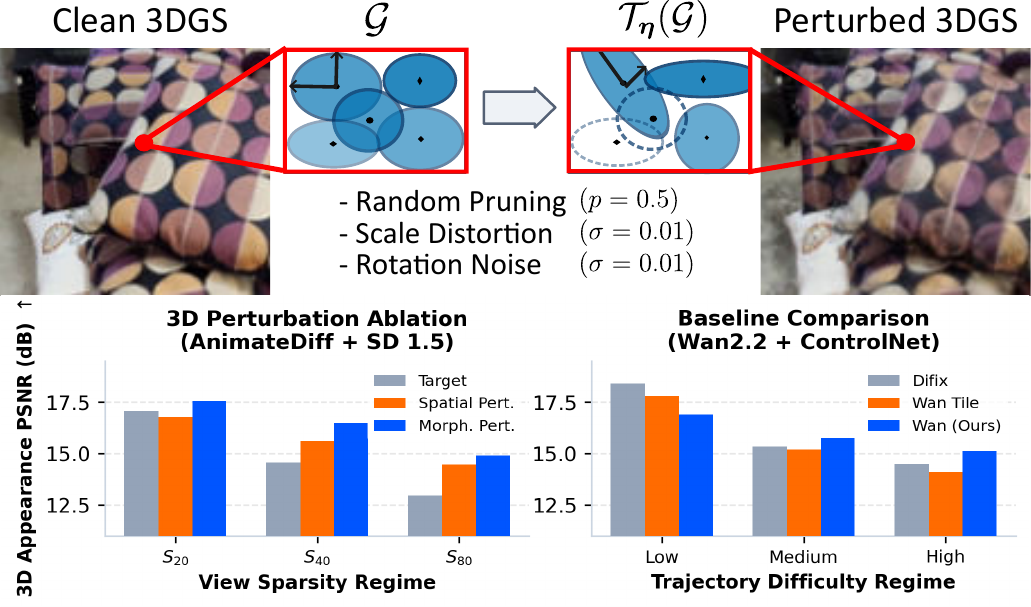}
  \caption{\textbf{3D Morphological Perturbations:} (Top) An optimization-free data augmentation generating spatially consistent paired renders without per-scene optimization. (Bottom) Training with our perturbed dataset instills superior geometric priors compared to sparse-view reconstructed pair sets. Adapting Wan \cite{wan2025wan} via ControlNet \cite{zhang2023adding} using our dataset outperforms general video refiners and state-of-the-art image artifact fixers.}
  \label{fig:teaser}
\end{figure}

Recent advances in 3D scene representations—most notably Neural Radiance Fields (NeRF) \cite{mildenhall2020nerf} and 3D Gaussian Splatting (3DGS) \cite{kerbl3Dgaussians}—have significantly improved the efficiency and fidelity of novel-view synthesis and scene reconstruction. These gains have advanced downstream domains like VR/AR \cite{fovnerf, 10.1145/3641519.3657448, 11092736} and robotics \cite{9712211, 10.1109/TRO.2025.3552348, 10160842}. However, because these representations are designed to overfit to dense input views, they struggle to generalize to views unobserved during training. This often results in inconsistent geometry, structural distortions, and rendering artifacts, especially in sparse-view settings.

To address the sparse-view limitation, recent research broadly follows two paradigms \cite{he2026survey, chen2026survey}. \textit{Regularization-based} methods inject explicit constraints—such as depth priors and cost volumes—into per-scene optimization \cite{li2024dngaussian, zhu2024fsgs, chen2024mvsplat}. Conversely, \textit{generalization-based} methods leverage generative priors by incorporating them into reconstruction pipelines or fine-tuning feed-forward architectures to directly regress Gaussian properties \cite{charatan2024pixelsplat, szymanowicz2024splatter, szymanowicz2025flash3d, xu2024grm}. Yet, explicit regularization degrades under extreme view sparsity, while feed-forward regression struggles with high-frequency details. Building on this, a new state-of-the-art paradigm—spearheaded by DiFix3D+ \cite{wu2025difix3d+}—fine-tunes large image or video diffusion backbones specifically to clean up 3DGS artifacts and hallucinate missing geometry across scene trajectories \cite{wu2025genfusion, 11176446, fischer2025flowr, de2026artifixer}. Despite their impressive results, these generative refinement models introduce a severe trade-off: 3DGS-specific artifact cleaning requires training on massive paired datasets of corrupted-scene renders and clean ground-truth views. To curate such datasets, existing approaches perform repeated specialized 3DGS optimizations, such as sparse-view \cite{wu2025difix3d+} or masked \cite{wu2025genfusion} reconstructions—a workflow that is computationally prohibitive to scale.

This bottleneck highlights a fundamental gap in simulating noise in 3D representations. In standard 2D computer vision, low-cost augmentations—such as Gaussian blur, color jitter, and random cropping—can be injected on-the-fly to implicitly regularize models \cite{shorten2019survey, cubuk2020randaugment}. NeRFLiX \cite{zhou2023nerflix} has used these to simulate NeRF reconstruction noise. However, no equivalent, optimization-free augmentations exist for 3D representations to obtain multi-view consistent render pairs: 2D frame-level noise violates multi-view consistency, while global sequence transforms cannot simulate camera-dependent scene artifacts.

To bridge this gap, we propose 3D Gaussian Perturbations as an optimization-free, 3D-aware regularizer. Our core idea is to leverage the explicit nature of 3D Gaussian Splatting, treating each Gaussian as a primitive unit of a scene, analogous to image pixels. Similar to how image data augmentations randomly perturb pixel values, we inject randomized noise directly into each Gaussian's underlying parameters. Applying perturbations directly to clean 3D Gaussians yields a corrupted scene that maintains epipolar consistency, enabling curation of corrupted-clean trajectory videos without per-scene 3DGS optimizations. Prior work \cite{hicom, recongs} has perturbed 3D Gaussian positions during scene reconstruction for stability, and we expand this idea to the full parameter space of 3D Gaussians (position $(\boldsymbol{\mu}_i)$, orientation $(\mathbf{q}_i)$, scale $(\mathbf{s}_i)$, color $(\mathbf{c}_i)$, and opacity $(\alpha_i)$). Through controlled diagnostic ablations in a lightweight video diffusion sandbox \cite{ICLR2024_94894cf9}, we isolate the effects of various combinations of parameter perturbations on motion module fine-tuning. We define \textbf{3D Morphological Perturbations} (Fig. 1)—scale ($s_i$) and orientation ($q_i$) perturbations combined with random pruning—as an optimal regularizer for 3D awareness. This approach stabilizes training dynamics and enables the model to outperform baselines trained on matching sparse-view distributions.

To assess the scalability and practical utility of our regularizer, we evaluate trajectory refinement on noisy 3DGS scenes generated via sparse-view reconstruction. Specifically, we train a ControlNet \cite{zhang2023adding} adapter for a 14-billion-parameter video diffusion backbone \cite{wan2025wan} using morphologically perturbed 3DGS data. Our model consistently outperforms standard Tile ControlNet baselines. Furthermore, it outperforms the Difix \cite{wu2025difix3d+} view-refiner model—reducing mean depth error by 12.5\% while preserving visual fidelity under extreme view sparsity (Fig. \ref{fig:teaser}). Finally, integrating our refinement model into a 3DGS-based scenario generator \cite{barcellona2025dream} translates these geometric gains into robotics impact, boosting imitation-learning policy success rates by up to 8.0\% across 3 of 4 manipulation tasks (e.g., jar closing). We summarize our main contributions as:
\begin{itemize}
    \item We propose 3D Gaussian Perturbations, an optimization-free data augmentation method that directly noise-corrupts 3DGS primitives—analogous to 2D image blur—to generate spatially consistent training pairs without per-scene optimization loops.
    \item We isolate \textbf{3D Morphological Perturbations} (scale, rotation, and primitive pruning) as a robust 3D regularizer, proving in video diffusion ablations that they yield stronger geometric priors than baselines trained on sparse-view reconstructions.
    \item We demonstrate scalability on a 14B video model via ControlNet adaptation, outperforming state-of-the-art artifact refiners with a 12.5\% mean depth error reduction that translates to up to an 8.0\% boost in downstream robotics policy success across tested manipulation tasks.
\end{itemize} 
\section{Related Works}
\label{sec:related_works}

\paragraph{Sparse-View 3DGS \& Generative Refinement.}
Overcoming novel-view synthesis artifacts under sparse inputs in representations like NeRF \cite{mildenhall2020nerf} and 3D Gaussian Splatting (3DGS) \cite{kerbl3Dgaussians} generally follows two paradigms \cite{he2026survey, chen2026survey}: \textit{regularization-based} methods that inject monocular depth or cost-volume constraints into per-scene optimization \cite{li2024dngaussian, zhu2024fsgs, chen2024mvsplat}, and \textit{generalization-based} approaches that feed-forward regress 3D primitives \cite{charatan2024pixelsplat, szymanowicz2024splatter, szymanowicz2025flash3d, xu2024grm}. To recover high-frequency details lost in feed-forward prediction, recent works leverage 2D/3D generative diffusion priors for view and trajectory refinement. Early refiners applied image-to-image models conditioned on reference views \cite{zhou2023nerflix, 10361604} or combined 2D refiners within iterative 3D update loops \cite{10.1007/978-3-031-72640-8_19, Paliwal_2025_ICCV, gsfix3d}. State-of-the-art frameworks—such as Difix3D+ \cite{wu2025difix3d+, Wu_2025_CVPR}, GenFusion \cite{wu2025genfusion, Wu_2025_CVPR}, 3DGS-Enhancer \cite{NEURIPS2024_f0b42291}, LM-Gaussian \cite{Yu2024LMGaussianBS, yu2025lmgaussianboostsparseview3d}, GuidedVD \cite{Zhong_2025_CVPR}, and GSFixer \cite{yin2025gsfixer}—fine-tune image or video diffusion backbones (e.g., CogVideoX \cite{yang2025cogvideox}) using spatial/feature visual conditioning \cite{wang2025vggt, oquab2024dinov} to resolve rendering corruptions \cite{11176446, fischer2025flowr, de2026artifixer}. However, training these generative models introduces a major bottleneck: curating massive paired corrupt-and-clean trajectory datasets requires repeated, optimization-heavy sparse-view or masked 3D reconstructions across every scene in the dataset. Our approach bypasses this costly curation pipeline entirely by synthesizing 3D-consistent training noise without per-scene optimization.
\vspace{-10pt}
\paragraph{Data Perturbations as Implicit Regularization}
Stochastic noise injection provides foundational implicit regularization: Bishop \cite{bishop1995training} demonstrated that zero-mean input noise is locally equivalent to an explicit $L_2$ gradient penalty, a principle generalized to feature and image spaces \cite{wager2013dropout, maaten2013learning, shorten2019survey}. In 3D vision, AugNeRF \cite{augnerf} perturbs input coordinates and output features to enhance NeRF robustness, while NeRFLiX \cite{zhou2023nerflix} applies image augmentations on ground-truth images to avoid expensive data collection for novel view refinement. Crucially, 3DGS representations are uniquely sensitive to parameter shifts, where small perturbations can cause severe degradation \cite{unified}. While prior works perturb only Gaussian positions ($\boldsymbol{\mu}_i$) to stabilize per-scene reconstruction \cite{hicom, recongs}, we generalize noise injection across all Gaussian parameters ($\mathbf{s}_i, \mathbf{q}_i, \mathbf{c}_i, \alpha_i$) as an optimization-free data augmentation for 3D-aware video diffusion fine-tuning.

\begin{comment}
\section{Preliminary: 3D Gaussian Splatting (3DGS)}
\label{sec:background}
In 3DGS \cite{kerbl3Dgaussians}, a scene is parameterized as an explicit collection of $M$ anisotropic primitives $\mathcal{G} = \{G_i\}_{i=1}^{M}$, defined by:
\begin{equation}
G_i = \{\boldsymbol{\mu}_i, \mathbf{q}_i, \mathbf{s}_i, \mathbf{c}_i, \alpha_i\},
\end{equation}
where $\boldsymbol{\mu}_i \in \mathbb{R}^3$ is the center, $\mathbf{q}_i \in \mathbb{S}^3$ is a rotation quaternion, $\mathbf{s}_i \in \mathbb{R}^3$ denotes scale, $\mathbf{c}_i \in \mathbb{R}^K$ represents color coefficients, and $\alpha_i \in [0,1]$ is opacity. Rendering a frame $I_k = \mathcal{R}(\mathcal{G}, \pi_k)$ from camera pose $\pi_k$ projects primitives into 2D screen space for point-wise $\alpha$-blending: $C(x) = \sum_{i \in N} \mathbf{c}_i \alpha_i' \prod_{j=1}^{i-1} (1 - \alpha_j')$, where $\alpha_i'$ is the projected 2D density scaled by $\alpha_i$.
\end{comment}
\section{Methodology}
\label{sec:methodology}
To train 3D-aware generative video models without the bottleneck of running an expensive data curation process with repeated scene reconstructions to obtain corrupted-clean video pairs, we introduce \textbf{3D Morphological Perturbations}—an optimization-free regularizer that injects stochastic noise directly into 3D primitive parameters to synthesize immediate, 3D-consistent scene degradations. In this section, we first briefly review 3D Gaussian Splatting, present the theoretical motivation behind 3D-aware perturbations, and detail our explicit parameter perturbation operations.
\vspace{-10pt}
\paragraph{Preliminary: 3D Gaussian Splatting.}
%\label{sec:background}
3DGS \cite{kerbl3Dgaussians} models a scene using $M$ primitives $\mathcal{G} = \{G_i\}_{i=1}^{M}$ defined by $G_i = \{\boldsymbol{\mu}_i, \mathbf{q}_i, \mathbf{s}_i, \mathbf{c}_i, \alpha_i\}$ (position $\boldsymbol{\mu}_i \in \mathbb{R}^3$, rotation $\mathbf{q}_i \in \mathbb{S}^3$, scale $\mathbf{s}_i \in \mathbb{R}^3$, color $\mathbf{c}_i \in \mathbb{R}^K$, and opacity $\alpha_i \in [0,1]$). Differentiable rasterization $I_k = \mathcal{R}(\mathcal{G}, \pi_k)$ projects primitives onto 2D screens for volumetric $\alpha$-blending.

% ===================================================================
% 3.1 THEORETICAL MOTIVATION
% ===================================================================

\subsection{3D Gaussian Perturbations as Regularization}
\label{sec:theory_motivation}
Our formulation builds upon the theoretical foundation of using stochastic noise injection to induce implicit regularization. Bishop \cite{bishop1995training} originally established that adding zero-mean noise is locally equivalent to training with an explicit input gradient penalty, a concept later generalized to feature-space corruptions \cite{maaten2013learning, wager2013dropout}. While these principles are widely applied in 2D image regularization \cite{shorten2019survey} and NeRF noise simulation \cite{zhou2023nerflix}, optimization-free augmentations in 3D space remain unexplored. Unlike prior work that perturbs only Gaussian positions during optimization for stability \cite{hicom, recongs}, we extend this analytical framework to the full parameter space of 3D Gaussians.

Consider a generative video refinement model $f_\theta(z_t, t, V)$ predicting noise residuals, conditioned on a rendered trajectory sequence $V = \mathcal{R}(\mathcal{G}, \Pi)$, where $\mathcal{R}$ represents the differentiable 3DGS rasterizer, $\mathcal{G}$ denotes the scene primitives, and $\Pi$ specifies the camera path. In standard 2D data augmentation, pixel-space perturbations $\tilde{V} = V + \boldsymbol{\delta}$ are injected directly into conditional inputs. In contrast, our approach injects stochastic perturbations $\boldsymbol{\eta} \sim p(\boldsymbol{\eta})$ directly into the primitive space $\mathcal{G}$, yielding perturbed renders $V(\boldsymbol{\eta}) = \mathcal{R}(\mathcal{T}_{\boldsymbol{\eta}}(\mathcal{G}), \Pi)$.

Assuming small noise $\boldsymbol{\eta}$, we can approximate the rasterizer $\mathcal{R}$ around uncorrupted parameters $\mathcal{G}$ using a first-order Taylor expansion:
\begin{equation}
V(\boldsymbol{\eta}) \approx V_0 + \mathbf{J}_{\mathcal{R}} \boldsymbol{\eta},
\end{equation}
where $V_0 = \mathcal{R}(\mathcal{G}, \Pi)$ is the clean render sequence, and $\mathbf{J}_{\mathcal{R}} = \frac{\partial \mathcal{R}}{\partial \mathcal{G}}$ is the rasterizer Jacobian at $\mathcal{G}$ (valid outside sorting thresholds).

Let $e_0 = f_\theta(z_t, t, V_0) - \epsilon$ be the clean diffusion prediction error. We similarly expand network output $f_\theta(z_t, t, V(\boldsymbol{\eta}))$ under the perturbed render condition:
\begin{equation}
f_\theta(z_t, t, V(\boldsymbol{\eta})) \approx f_\theta(z_t, t, V_0) + \mathbf{J}_{f} \mathbf{J}_{\mathcal{R}} \boldsymbol{\eta},
\end{equation}
where $\mathbf{J}_{f} = \frac{\partial f_\theta}{\partial V_0}$ measures model sensitivity to conditional input variations. Under this local linear approximation, training $f_\theta$ on 3D-perturbed renders optimizes an empirical surrogate for condition sensitivity:
\begin{equation}
\label{eq:relaxed_regularizer}
\begin{aligned}
\mathcal{L}_{\text{pert}}(\theta) &= \mathbb{E}_{\boldsymbol{\eta}} \left[ \left\| f_\theta(z_t, t, V(\boldsymbol{\eta})) - \epsilon \right\|_2^2 \right] \\
&\approx \mathbb{E}_{\boldsymbol{\eta}} \left[ \left\| e_0 + \mathbf{J}_{f} \mathbf{J}_{\mathcal{R}} \boldsymbol{\eta} \right\|_2^2 \right] \approx \mathcal{L}_{\text{clean}}(\theta) + \mathcal{R}_{\text{manifold}}(\theta),
\end{aligned}
\end{equation}
where $\mathcal{L}_{\text{clean}}(\theta) = \mathbb{E}\left[\| e_0 \|_2^2\right]$ is the baseline noise-matching objective, and $\mathcal{R}_{\text{manifold}}(\theta) \propto \mathbb{E}_{\boldsymbol{\eta}} \left[ \left\| \mathbf{J}_{f} \mathbf{J}_{\mathcal{R}} \boldsymbol{\eta} \right\|_2^2 \right]$ implicitly penalizes local predictions along the rendering manifold.

Eq.~\ref{eq:relaxed_regularizer} shows that 3D primitive perturbations act as an implicit, manifold-constrained regularizer \cite{bishop1995training}. Projecting network sensitivity ($\mathbf{J}_f$) through the rasterizer Jacobian ($\mathbf{J}_{\mathcal{R}}$) ensures that regularization specifically targets geometry-consistent surface variations, rather than random 2D pixel noise. This trains the model to actively resolve 3D structural artifacts while maintaining strict multi-view consistency. Non-Gaussian operations like pruning build on this, forcing the network to recover missing geometry.

% ===================================================================
% 3.2 3D GAUSSIAN PERTURBATIONS
% ===================================================================

\subsection{Perturbation Categories}
\label{sec:perturbations}

\begin{figure}[t]
  \centering
  \includegraphics[width=\columnwidth]{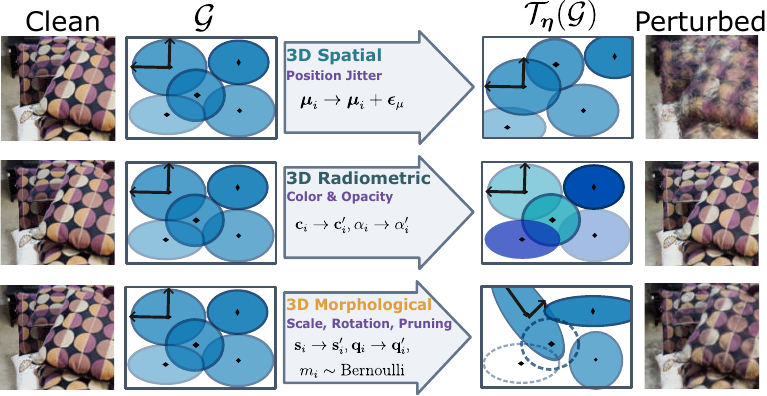}
  \caption{Overview of proposed 3D primitive perturbations. \textbf{Spatial} noise perturbs primitive centers ($\boldsymbol{\mu}$); \textbf{Radiometric} noise alters appearance ($\mathbf{c}, \alpha$); \textbf{Morphological} perturbations modify local geometry ($\mathbf{s}, \mathbf{q}$) and structural density ($m$) to simulate 3D scene degradations.}
  \label{fig:method}
  \vspace{-10pt}
\end{figure}

To implement the regularizer derived in Section~\ref{sec:theory_motivation}, we translate the theoretical noise vector $\boldsymbol{\eta}$ into concrete parameter operations $\mathcal{T}_{\boldsymbol{\eta}}(\mathcal{G})$ applied to each primitive $G_i = \{\boldsymbol{\mu}_i, \mathbf{q}_i, \mathbf{s}_i, \mathbf{c}_i, \alpha_i\}$. Rather than adding random noise everywhere, we divide these operations into three distinct categories—\textit{Spatial}, \textit{Radiometric}, and \textit{Morphological} (Figure~\ref{fig:method}). Each category targets specific primitive attributes to model common 3D reconstruction artifacts.

\subsubsection{3D Spatial Perturbations}
Inspired by \cite{hicom, recongs}, we isolate physical translation noise across primitive centers:

\vspace{-10pt}
\paragraph{Position Jitter.}
To account for spatial coordinate drift independently of scene scale, we perturb primitive centers $\boldsymbol{\mu}_i$ proportional to the bounding box extent $E_{\text{avg}} = \frac{1}{3} \sum_{d \in \{x,y,z\}} (\max \boldsymbol{\mu}_{\cdot, d} - \min \boldsymbol{\mu}_{\cdot, d})$:
\vspace{-5pt}
\begin{equation}
\boldsymbol{\mu}_i' = \boldsymbol{\mu}_i + \boldsymbol{\epsilon}_\mu, \quad \boldsymbol{\epsilon}_\mu \sim \mathcal{N}\left(\mathbf{0}, (\gamma_{\text{xyz}} \cdot E_{\text{avg}})^2 \mathbf{I}_3\right),
\end{equation}
%\vspace{-3pt}
where $\gamma_{\text{xyz}} = 0.002$ scales the spatial jitter relative to the scene domain.

\subsubsection{3D Radiometric Perturbations}
Complementing spatial position jitter, we isolate non-geometric appearance noise across primitive color and opacity parameters:

\vspace{-10pt}
\paragraph{Photometric Noise.}
To approximate illumination inconsistencies and sensor noise across views, additive zero-mean Gaussian noise is applied to the RGB color coefficients $\mathbf{c}_i$, clamped to the valid range $[0, 1]$:
\vspace{-5pt}
\begin{equation}
\mathbf{c}_i' = \operatorname{clamp}\left(\mathbf{c}_i + \boldsymbol{\epsilon}_c, \, 0, \, 1\right), \quad \boldsymbol{\epsilon}_c \sim \mathcal{N}(\mathbf{0}, \sigma_{\text{color}}^2 \mathbf{I}_3),
\end{equation}
where $\sigma_{\text{color}} = 0.05$ bounds color noise to $5\%$ of the normalized range.

\vspace{-10pt}
\paragraph{Opacity Noise.}
Density estimation errors and semi-transparent rendering artifacts are modeled by perturbing primitive opacities $\alpha_i$, bounded strictly away from 0 and 1 to preserve numerical stability during rasterization:
\vspace{-5pt}
\begin{equation}
\alpha_i' = \operatorname{clamp}\left(\alpha_i + \epsilon_\alpha, \, \epsilon_0, \, 1 - \epsilon_0\right), \quad \epsilon_\alpha \sim \mathcal{N}(0, \sigma_{\text{opac}}^2),
\end{equation}
where $\sigma_{\text{opac}} = 0.05$ and $\epsilon_0 = 10^{-4}$.
%\vspace{-10pt}
\subsubsection{3D Morphological Perturbations}
Unlike position-only spatial perturbations, we propose 3D morphological perturbations to alter primitive extents and density, simulating structural degradation and volume loss while preserving global alignment:

\vspace{-10pt}
\paragraph{Random Pruning.}
Sparse-view reconstructions frequently exhibit missing geometry due to insufficient spatial coverage. To approximate this effect, we sample a binary retention mask $m_i \in \{0, 1\}$ for each primitive:
\vspace{-5pt}
\begin{equation}
m_i \sim \text{Bernoulli}(1 - p_{\text{prune}}), \quad \mathcal{G}' = \{G_i \in \mathcal{G} \mid m_i = 1\},
\end{equation}
where $p_{\text{prune}}$ is the drop probability. Setting $p_{\text{prune}} = 0.5$ reduces the expected scene density by half ($\mathbb{E}[M'] = 0.5M$), forcing the model to reconstruct complete surfaces from sparse primitive coverage.

\vspace{-10pt}
\paragraph{Scale Perturbation.}
To model inaccurate Gaussian proportion distortions while maintaining positive scales, additive zero-mean Gaussian noise is applied to each coordinate component $j \in \{1, 2, 3\}$:
\vspace{-5pt}
\begin{equation}
s_{i, j}' = \max\left(s_{i, j} + \epsilon_{s, j}, \, 10^{-6}\right), \quad \epsilon_{s, j} \sim \mathcal{N}(0, \sigma_{\text{scale}}^2),
\end{equation}
where $\sigma_{\text{scale}} = 0.01$ models $1\%$ anisotropic scale jitter, simulating shape distortion without degenerate primitives.

\vspace{-10pt}
\paragraph{Rotation Perturbation.}
Local orientation misalignment is modeled by injecting element-wise Gaussian noise directly onto the quaternion coefficients followed by $\ell_2$-renormalization to maintain $\mathbf{q}_i' \in \mathbb{S}^3$:
\vspace{-5pt}
\begin{equation}
\mathbf{q}_i' = \frac{\mathbf{q}_i + \boldsymbol{\epsilon}_q}{\|\mathbf{q}_i + \boldsymbol{\epsilon}_q\|_2}, \quad \boldsymbol{\epsilon}_q \sim \mathcal{N}(\mathbf{0}, \sigma_{\text{rot}}^2 \mathbf{I}_4),
\end{equation}
where $\sigma_{\text{rot}} = 0.01$ introduces minor orientation drift ($\approx 1.15^\circ$ angular jitter) to model surface normal misalignment while preserving local geometry.

Injecting perturbations directly into 3D primitive space guarantees that rendering trajectory sequences $\mathcal{R}(\mathcal{T}_{\boldsymbol{\eta}}(\mathcal{G}), \pi_k)$ across camera path $\Pi = \{\pi_k\}_{k=1}^K$ maintain exact multi-view epipolar consistency, providing a regularizing signal without requiring optimization.
\section{Experiments}
\label{sec:experiments}

\begin{figure*}[htbp]
  \centering
  \includegraphics[width=\linewidth]{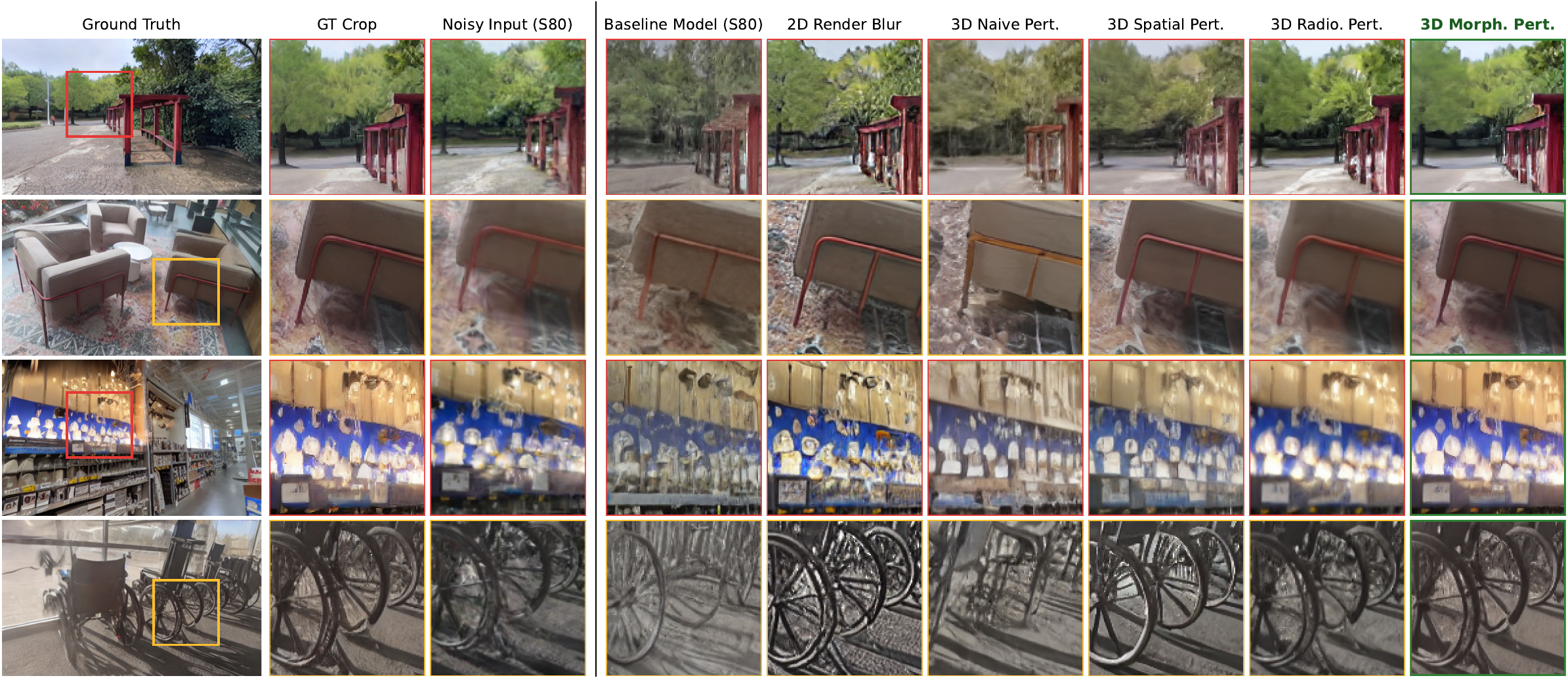}
  \caption{\textbf{Qualitative comparison of 3D scenes reconstructed from corrupted video trajectories refined by diffusion models trained on each perturbed dataset.} Evaluated on $\mathcal{S}_{80}$ test set using 3DGS representations re-optimized from refined output frames. \textbf{3D Morphological Perturbation} best recovers fine geometry and appearance fidelity relative to ground truth.}
  \label{fig:animatediff_quantitative}
\end{figure*}

\begin{table*}[t]
\centering
\scriptsize
\renewcommand{\arraystretch}{0.90}
%\captionsetup{font=scriptsize}
\caption{\textbf{Ablation of 3D perturbation strategies across sparse-view regimes ($\mathcal{S}_{20}$, $\mathcal{S}_{40}$, and $\mathcal{S}_{80}$).} Perturbation strategies are applied to clean 3DGS scenes during training dataset preparation, and a separate instance of our lightweight video diffusion model is trained on each resulting subset to isolate its effect. Evaluated on held-out test scenes across 2D Video Refinement ($\text{PSNR}_{\text{2D}}$, $\text{SSIM}_{\text{2D}}$, $\text{LPIPS}_{\text{2D}}$, $E_{\text{flow}}$), 3DGS Appearance ($\text{PSNR}_{\text{3D}}$, $\text{SSIM}_{\text{3D}}$, $\text{LPIPS}_{\text{3D}}$), and 3DGS Geometry ($\text{CD}$, $\text{F-Score}$). Best scores are in \textbf{bold}; second-best in \textit{italics}.}
\label{tab:main_sparse_results_compact}
\begin{tabular*}{1.0\textwidth}{@{\extracolsep{\fill}} l | l | cccc | ccc | cc }
\toprule
\textbf{Regime} & \textbf{Training Method} & \textbf{PSNR}$_{\text{2D}}\uparrow$ & \textbf{SSIM}$_{\text{2D}}\uparrow$ & \textbf{LPIPS}$_{\text{2D}}\downarrow$ & \textbf{$E_{\text{flow}}$}$\downarrow$ & \textbf{PSNR}$_{\text{3D}}\uparrow$ & \textbf{SSIM}$_{\text{3D}}\uparrow$ & \textbf{LPIPS}$_{\text{3D}}\downarrow$ & \textbf{CD}$\downarrow$ & \textbf{F-Score}$\uparrow$ \\
\midrule
\multirow{6}{*}{\textbf{$S_{20}$}} 
& $\star$ Stride 20 (Target) & 17.018 & 0.572 & 0.412 & 0.252 & 17.074 & 0.594 & 0.392 & \cellcolor{green!20}{\textbf{0.009}} & 0.771 \\
& 2D Render Blur & 16.590 & 0.490 & 0.431 & 0.235 & 16.855 & 0.516 & 0.412 & 0.010 & 0.751 \\
& 3D Naive Pert. & 15.100 & 0.375 & 0.554 & 0.311 & 15.193 & 0.407 & 0.543 & 0.011 & 0.730 \\
& 3D Spatial Pert. & 16.629 & 0.524 & 0.441 & 0.250 & 16.761 & 0.549 & 0.419 & 0.010 & 0.759 \\
& 3D Radiometric Pert. & \cellcolor{green!20}{\textbf{17.907}} & \cellcolor{green!20}{\textbf{0.621}} & \cellcolor{green!20}{\textbf{0.373}} & \cellcolor{green!20}{\textbf{0.219}} & \cellcolor{green!20}{\textbf{17.974}} & \cellcolor{green!20}{\textbf{0.635}} & \cellcolor{green!20}{\textbf{0.357}} & \cellcolor{green!20}{\textbf{0.009}} & \cellcolor{green!10}{\textit{0.781}} \\
& \textbf{3D Morphological Pert.} & \cellcolor{green!10}{\textit{17.487}} & \cellcolor{green!10}{\textit{0.601}} & \cellcolor{green!10}{\textit{0.397}} & \cellcolor{green!10}{\textit{0.229}} & \cellcolor{green!10}{\textit{17.546}} & \cellcolor{green!10}{\textit{0.616}} & \cellcolor{green!10}{\textit{0.379}} & \cellcolor{green!20}{\textbf{0.009}} & \cellcolor{green!20}{\textbf{0.786}} \\
\cmidrule(lr){1-11}
\multirow{6}{*}{\textbf{$S_{40}$}} 
& $\star$ Stride 40 (Target) & 14.503 & 0.428 & 0.525 & 0.331 & 14.552 & 0.454 & 0.508 & 0.011 & 0.733 \\
& 2D Render Blur & 15.439 & 0.436 & 0.481 & 0.271 & 15.652 & 0.462 & 0.463 & 0.010 & 0.736 \\
& 3D Naive Pert. & 14.487 & 0.338 & 0.582 & 0.333 & 14.547 & 0.370 & 0.573 & 0.011 & 0.719 \\
& 3D Spatial Pert. & 15.548 & 0.476 & 0.484 & 0.283 & 15.616 & 0.500 & 0.464 & 0.010 & 0.749 \\
& 3D Radiometric Pert. & \cellcolor{green!10}{\textit{16.366}} & \cellcolor{green!20}{\textbf{0.560}} & \cellcolor{green!20}{\textbf{0.431}} & \cellcolor{green!20}{\textbf{0.254}} & \cellcolor{green!10}{\textit{16.365}} & \cellcolor{green!20}{\textbf{0.574}} & \cellcolor{green!20}{\textbf{0.415}} & \cellcolor{green!20}{\textbf{0.009}} & \cellcolor{green!10}{\textit{0.768}} \\
& \textbf{3D Morphological Pert.} & \cellcolor{green!20}{\textbf{16.479}} & \cellcolor{green!10}{\textit{0.552}} & \cellcolor{green!10}{\textit{0.445}} & \cellcolor{green!10}{\textit{0.257}} & \cellcolor{green!20}{\textbf{16.483}} & \cellcolor{green!10}{\textit{0.567}} & \cellcolor{green!10}{\textit{0.426}} & \cellcolor{green!20}{\textbf{0.009}} & \cellcolor{green!20}{\textbf{0.774}} \\
\cmidrule(lr){1-11}
\multirow{6}{*}{\textbf{$S_{80}$}} 
& $\star$ Stride 80 (Target) & 12.905 & 0.286 & 0.613 & 0.392 & 12.961 & 0.315 & 0.609 & 0.013 & 0.690 \\
& 2D Render Blur & 14.103 & 0.389 & 0.528 & 0.312 & 14.260 & 0.414 & 0.510 & 0.011 & 0.724 \\
& 3D Naive Pert. & 13.788 & 0.315 & 0.605 & 0.357 & 13.819 & 0.346 & 0.597 & 0.012 & 0.710 \\
& 3D Spatial Pert. & 14.449 & 0.433 & 0.528 & 0.311 & 14.460 & 0.454 & 0.510 & 0.011 & 0.736 \\
& 3D Radiometric Pert. & \cellcolor{green!10}{\textit{14.761}} & \cellcolor{green!10}{\textit{0.497}} & \cellcolor{green!20}{\textbf{0.491}} & \cellcolor{green!20}{\textbf{0.298}} & \cellcolor{green!10}{\textit{14.712}} & \cellcolor{green!10}{\textit{0.509}} & \cellcolor{green!20}{\textbf{0.475}} & \cellcolor{green!20}{\textbf{0.010}} & \cellcolor{green!10}{\textit{0.754}} \\
& \textbf{3D Morphological Pert.} & \cellcolor{green!20}{\textbf{14.936}} & \cellcolor{green!20}{\textbf{0.502}} & \cellcolor{green!10}{\textit{0.497}} & \cellcolor{green!20}{\textbf{0.298}} & \cellcolor{green!20}{\textbf{14.907}} & \cellcolor{green!20}{\textbf{0.517}} & \cellcolor{green!10}{\textit{0.481}} & \cellcolor{green!20}{\textbf{0.010}} & \cellcolor{green!20}{\textbf{0.762}} \\
\bottomrule
\end{tabular*}
\end{table*}

\begin{figure}[tbp]
  \centering
  \includegraphics[width=\columnwidth]{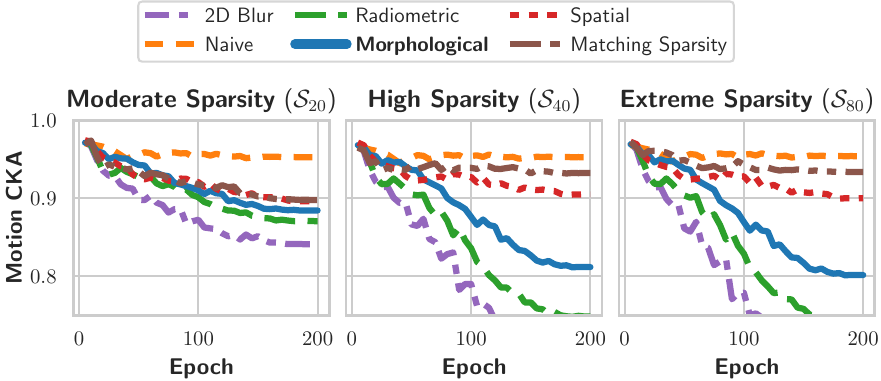}
  %\captionsetup{font=scriptsize}
  \caption{\textbf{Centered Kernel Analysis (CKA) similarity of AnimateDiff motion module features across models trained on different perturbed datasets.} Measured relative to initialization on temporal attention feature maps throughout training to evaluate representation stability and activation drift under different perturbation strategies.}
  \label{fig:cka_trajectory_ablation}
\end{figure}

We validate our perturbation framework across a three-stage experimental progression. First, Section~\ref{sec:ablation} conducts a diagnostic ablation using a lightweight video diffusion sandbox to isolate the effect of individual perturbation categories (Section~\ref{sec:perturbations}). Next, Section~\ref{sec:3dgtr} scales this proof of concept to a video foundation model (Wan2.2 \cite{wan2025wan}) via ControlNet \cite{zhang2023adding} adapters trained on our perturbed Gaussian data. Finally, Section~\ref{sec:robotics_downstream} evaluates the downstream application of our refined trajectories for robotic imitation learning within a 3DGS-based scenario generator~\cite{barcellona2025dream}.
%\vspace{-10pt}
\subsection{Diagnostic Ablation: Perturbation Categories}
\label{sec:ablation}

\paragraph{Model Architecture.}
We construct a lightweight video diffusion sandbox using a Stable Diffusion 1.5~\cite{rombach2021highresolution} UNet augmented with AnimateDiff~\cite{guo2024animatediff} temporal motion modules. To enable video-to-video refinement, the first convolutional layer is expanded to 8 input channels to ingest corrupted Gaussian conditioning latents $z_{\text{cond}}$. Fine-tuning is strictly restricted to rank-$r$ LoRA~\cite{hu2022lora} adapters injected into spatial and temporal cross-attention blocks while base weights remain frozen. Because neither pre-trained model possesses knowledge of 3D geometric representations, the architecture relies on $z_{\text{cond}}$ for all spatial alignment, ensuring that scene geometry originates strictly from our 3D Gaussian renderings. Further implementation details are included in \textit{Supplementary Material}.
\vspace{-10pt}
\paragraph{Datasets and Baselines.}
We randomly sample 250 scenes from DL3DV-10K large-scale scene dataset \cite{Ling_2024_CVPR} (200 train, 50 test). For training scenes, we reconstruct baseline 3D Gaussian Splatting \cite{kerbl3Dgaussians} representations and individually apply each perturbation category from Section~\ref{sec:methodology}: \textit{Morphological} (Pruning, Scale, Rotation), \textit{Spatial} (Jitter), and \textit{Radiometric} (Photometric, Opacity), alongside a \textit{Naive} baseline applying all types simultaneously. Corrupted 16-frame conditioning sequences rendered along smooth trajectories are paired with uncorrupted ground-truth renders. As a 2D baseline, a \textit{2D Render Blur} set is synthesized via spatial Gaussian smoothing on clean ground-truth trajectories. We train separate model instances on each subset to isolate their individual effects. Held-out test scenes are reconstructed under three sparse-view regimes: \textit{Moderate} ($\mathcal{S}_{20}$), \textit{High} ($\mathcal{S}_{40}$), and \textit{Extreme} ($\mathcal{S}_{80}$) sparsities, where $\mathcal{S}_k$ uses every $k$-th view from the training trajectory. Finally, we compare against model instances trained directly on matching sparse-view reconstructions.
\vspace{-10pt}

\paragraph{Metrics.} We evaluate across three levels: 2D Video (PSNR, SSIM, LPIPS, flow error $E_{\text{flow}}$); 3DGS Appearance (PSNR, SSIM, LPIPS on 2 held-out views per 16-frame render); and 3DGS Geometry (Chamfer Distance CD, F-Score). To monitor how internal representations evolve from initialization during training, we compute Centered Kernel Analysis (CKA)~\cite{pmlr-v97-kornblith19a} similarity matrices on temporal attention maps over training time.
\vspace{-6pt}
\paragraph{Results.}
As detailed in Table~\ref{tab:main_sparse_results_compact} and Figure~\ref{fig:animatediff_quantitative}, as view sparsity increases to $\mathcal{S}_{80}$, target baselines fail to generalize due to severe overfitting, whereas all perturbation strategies effectively mitigate overfitting. However, perturbation choice is critical: \textit{3D Naive Perturbation} corrupts all Gaussian parameters simultaneously, failing to reconstruct coherent geometry or color, while \textit{2D Render Blur} violates multi-view epipolar consistency and introduces non-photorealistic edge artifacts. Among isolated 3D perturbations, \textit{3D Spatial Perturbation} recovers sharp geometry with color degradation, whereas \textit{3D Radiometric Perturbation} preserves appearance but degrades in geometric fidelity ($\text{CD}$, $\text{F-Score}$) under extreme sparsity ($\mathcal{S}_{80}$). In contrast, \textbf{3D Morphological Perturbation} achieves high geometric fidelity and appearance most accurate to ground truth, leading the model to infer missing structure rather than rely on color cues.

Examining temporal attention feature dynamics in Figure~\ref{fig:cka_trajectory_ablation} further clarifies these optimization trends. When training our AnimateDiff-based model, \textit{3D Naive Perturbation} fails to drive meaningful feature learning, leaving activations stagnant at $\sim 95\%$ similarity to initialization. In contrast, \textit{2D Render Blur} excessively alters internal activations, causing feature trajectories to diverge drastically from target baselines. The isolated 3D perturbations display the most consistent CKA curves, even surpassing the stability of matching baselines. Among them, \textbf{3D Morphological Perturbation} results in the cleanest curve with the smallest fluctuations, while \textit{3D Spatial Perturbation} remains closest to the baseline feature dynamics. Given its robustness under extreme sparsity, dominant 3D geometric fidelity, and superior representation stability, we select \textbf{3D Morphological Perturbation} as the optimal strategy for scaling to large foundation models in Section~\ref{sec:3dgtr}. Additional results across all sparsity settings are included in \textit{Suppl. Material}.

\begin{figure*}[htbp]
  \centering
  \includegraphics[width=\linewidth]{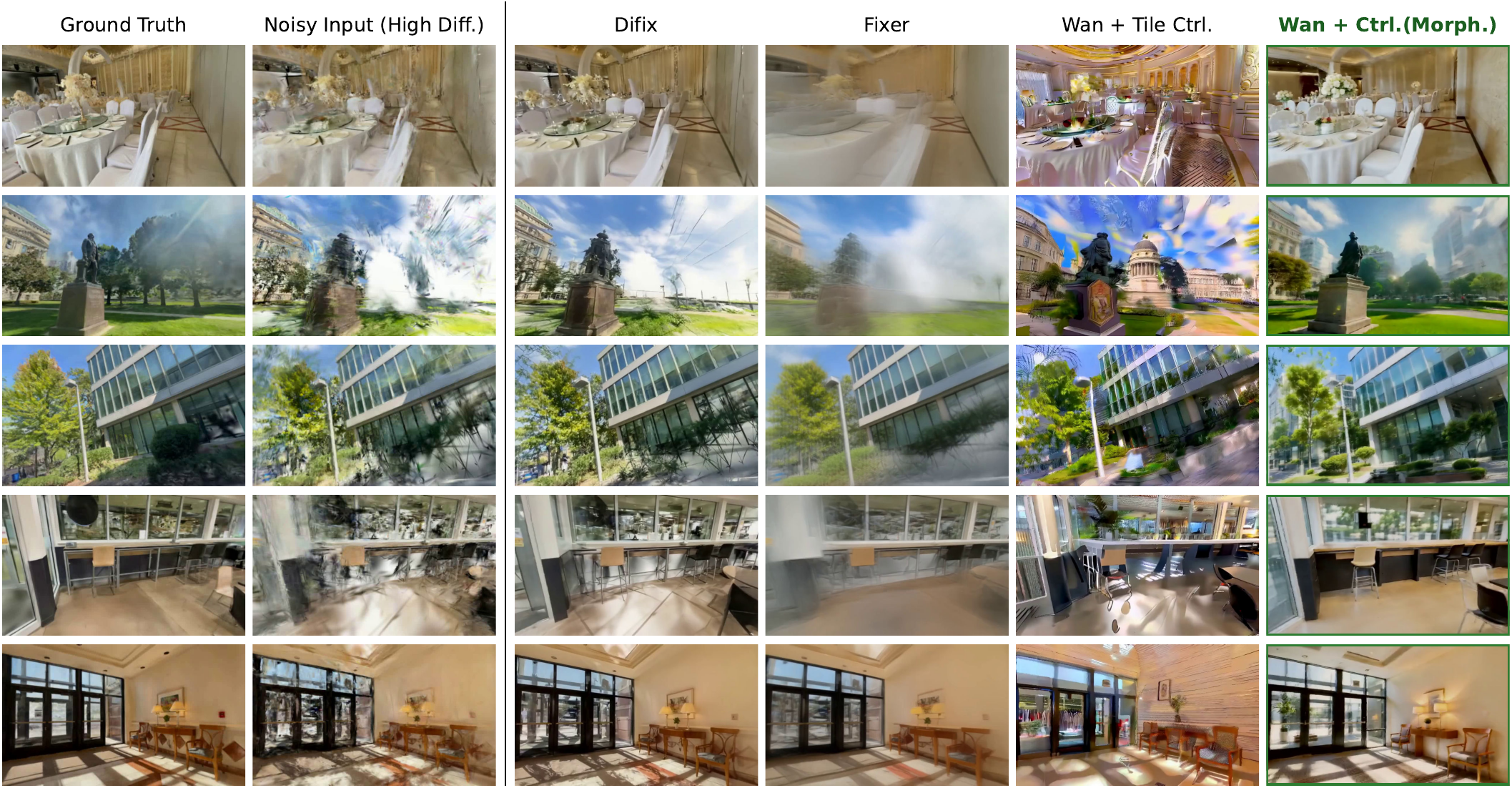}
  %\captionsetup{font=scriptsize}
  \caption{\textbf{Qualitative comparison on \textit{High difficulty} (see Sec. \ref{sec:3dgtr}) test scenes.} Frame-by-frame baselines either fail to refine artifacts (\textit{Difix}) or introduce severe blur (\textit{Fixer}). \textbf{Wan-Tile} hallucinated features that diverge from the original scene. Our \textbf{Wan + Ctrl (Morph.)} successfully removes reconstruction artifacts and inpaints seamless, structure-preserving details that closely align with the true scene geometry.
\label{fig:qualitative_results}}
  \label{fig:wan_quantitative}
\end{figure*}

\begin{table*}[t]
\centering
%\captionsetup{font=scriptsize}
\caption{\textbf{Quantitative evaluation of trajectory-level video refinement on the DL3DV benchmark across varying view sparsity and pose overlap regimes.} Evaluated across 49-frame central trajectories rendered at $832\times480$. Standard RGB metrics evaluate appearance fidelity, while depth metrics assess geometric accuracy. Despite having a denser 6-view reconstruction, the \textit{High Difficulty} setting poses the hardest refinement task because its test trajectory is rendered entirely off-reference (pure pose interpolation). Best proxy scores are in \textbf{dark green}; second-best in \textit{light green}. (Wan2.2 + Contolnet scaled experiment)}
\label{tab:eccv_heatmap_views_subtle}
\scriptsize
\setlength{\tabcolsep}{3.5pt}
\renewcommand{\arraystretch}{1.12}
\begin{tabular*}{1.0\textwidth}{@{\extracolsep{\fill}} l | cccc | cccc | cccc }
\toprule
\textbf{Difficulty Regime} 
& \multicolumn{4}{c|}{\textbf{Low Difficulty}} 
& \multicolumn{4}{c|}{\textbf{Medium Difficulty}} 
& \multicolumn{4}{c}{\textbf{High Difficulty}} \\

\textbf{Train views / Test overlap} 
& \multicolumn{4}{c|}{\textit{9 Views / 1 View Overlapping w. Test Traj.}} 
& \multicolumn{4}{c|}{\textit{3 Views / 1 View Overlapping w. Test Traj.}} 
& \multicolumn{4}{c}{\textit{6 Views (No overlap w. Test Traj.)}} \\
\hline

\textbf{Metric}
& Difix & Fixer & \shortstack{Wan-Tile} & \shortstack{\textbf{Wan (Morph.)}} 
& Difix & Fixer & \shortstack{Wan-Tile} & \shortstack{\textbf{Wan (Morph.)}} 
& Difix & Fixer & \shortstack{Wan-Tile} & \shortstack{\textbf{Wan (Morph.)}} \\
\hline

PSNR$\uparrow$    
& 18.39 & \cellcolor{green!20}{\textbf{18.63}} & 14.48 & \cellcolor{green!10}\textit{16.90}
& 15.34 & \cellcolor{green!20}{\textbf{16.07}} & 13.56 & \cellcolor{green!10}\textit{15.76}
& 14.48 & \cellcolor{green!20}{\textbf{15.19}} & 13.23 & \cellcolor{green!10}\textit{15.11} \\

\rowcolor{gray!10}
SSIM$\uparrow$    
& 0.651 & \cellcolor{green!20}{\textbf{0.690}} & 0.394 & \cellcolor{green!10}\textit{0.610}
& 0.551 & \cellcolor{green!20}{\textbf{0.627}} & 0.379 & \cellcolor{green!10}\textit{0.570}
& 0.516 & \cellcolor{green!20}{\textbf{0.604}} & 0.365 & \cellcolor{green!10}\textit{0.549} \\

LPIPS$\downarrow$ 
& \cellcolor{green!20}{\textbf{0.336}} & \cellcolor{green!10}\textit{0.366} & 0.618 & 0.416
& \cellcolor{green!10}\textit{0.431} & \cellcolor{green!20}{\textbf{0.441}} & 0.640 & 0.461
& \cellcolor{green!20}{\textbf{0.492}} & 0.550 & 0.679 & \cellcolor{green!10}\textit{0.502} \\
\hline

\rowcolor{gray!10}
AbsRel$\downarrow$ 
& \cellcolor{green!20}{\textbf{0.241}} & 0.398 & 0.294 & \cellcolor{green!10}\textit{0.264}
& \cellcolor{green!10}\textit{0.322} & 0.438 & 0.363 & \cellcolor{green!20}{\textbf{0.299}}
& \cellcolor{green!10}\textit{0.308} & 0.555 & 0.360 & \cellcolor{green!20}{\textbf{0.306}} \\

RMSE$\downarrow$   
& \cellcolor{green!10}\textit{24.20} & 43.68 & 27.58 & \cellcolor{green!20}{\textbf{16.18}}
& \cellcolor{green!10}\textit{31.67} & 48.20 & 35.08 & \cellcolor{green!20}{\textbf{30.65}}
& \cellcolor{green!10}\textit{29.87} & 54.24 & 32.03 & \cellcolor{green!20}{\textbf{29.53}} \\

\rowcolor{gray!10}
$\delta_1$$\uparrow$ 
& \cellcolor{green!20}{\textbf{0.684}} & 0.564 & 0.626 & \cellcolor{green!10}\textit{0.657}
& \cellcolor{green!10}\textit{0.544} & 0.457 & 0.465 & \cellcolor{green!20}{\textbf{0.587}}
& \cellcolor{green!10}\textit{0.553} & 0.424 & 0.524 & \cellcolor{green!20}{\textbf{0.576}} \\

$\delta_2$$\uparrow$ 
& \cellcolor{green!20}{\textbf{0.868}} & 0.767 & 0.848 & \cellcolor{green!10}\textit{0.843}
& \cellcolor{green!10}\textit{0.783} & 0.700 & 0.766 & \cellcolor{green!20}{\textbf{0.794}}
& \cellcolor{green!10}\textit{0.806} & 0.627 & 0.787 & \cellcolor{green!20}{\textbf{0.821}} \\

\rowcolor{gray!10}
$\delta_3$$\uparrow$ 
& \cellcolor{green!20}{\textbf{0.928}} & 0.858 & 0.918 & \cellcolor{green!20}{\textbf{0.928}}
& \cellcolor{green!10}\textit{0.889} & 0.841 & 0.886 & \cellcolor{green!20}{\textbf{0.908}}
& \cellcolor{green!10}\textit{0.922} & 0.765 & 0.886 & \cellcolor{green!20}{\textbf{0.924}} \\
\bottomrule
\end{tabular*}
\end{table*}

\begin{figure}[htbp]%[th!]
      \centering
      \includegraphics[width=\columnwidth]{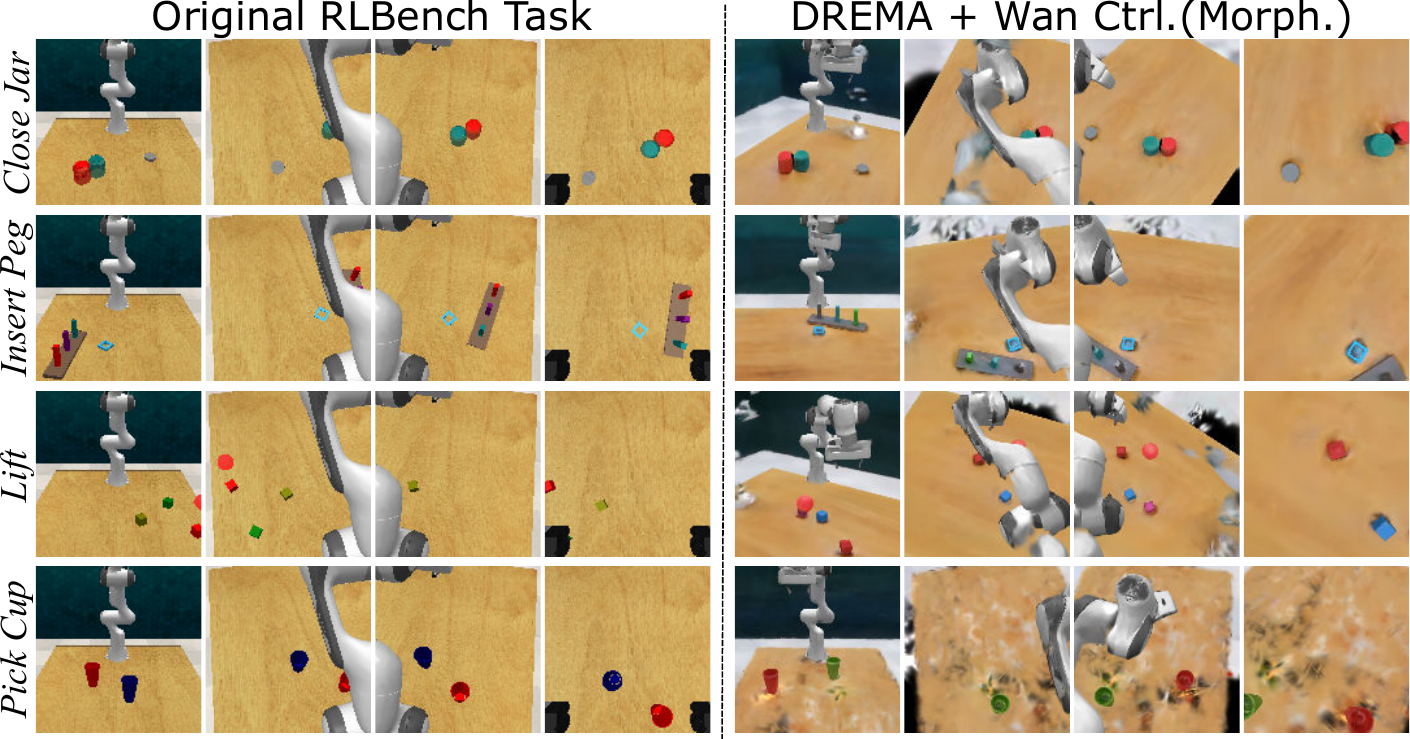}
      %\captionsetup{font=scriptsize}
      \caption{Qualitative comparison between original tasks from RLBench \cite{james2019rlbench} and tasks generated with DREMA + Wan Ctrl (Morph.) refinement. The reported four views: \textit{front}, \textit{left shoulder}, \textit{right shoulder} and \textit{wrist}.}
      \label{fig:qualitative_rlbench}
\end{figure}

\subsection{Scaling to Video Foundation Models for Trajectory Render Refinement}
\label{sec:3dgtr}
\paragraph{Model Architecture and Training.}
To scale our 3D refinement framework, we build upon Wan2.2~\cite{wan2025wan}, a DiT-based \cite{peebles2023scalable} text-to-video foundation model. We instantiate a dilated ControlNet~\cite{zhang2023adding} by copying 8 DiT blocks at a stride of 3 from the frozen Wan base model. Clean target frames are encoded using Wan2.2's VAE, while text captions generated via Qwen2~\cite{yang2024qwen2technicalreport} provide text conditioning. During the forward pass, the ControlNet processes corrupted 49-frame trajectory renderings ($832\times480$ resolution) and injects its intermediate activations additively into the corresponding frozen base DiT blocks. The model is trained for 30 epochs across 8 GPUs with 80GB VRAM. Further implementation details are included in \textit{Supplementary Material}.

\paragraph{Dataset Curation and Training.}
We train the ControlNet on ~10K perturbed-clean 49-frame video pairs aggregated across DL3DV-10K~\cite{Ling_2024_CVPR} ($75\%$), ScanNet~\cite{dai2017scannet} ($15\%$), and ScanNet++~\cite{yeshwanth2023scannet++} ($10\%$) ($9,866$ pairs total). Each scene is perturbed with our 3D Morphological Perturbations (Sec. \ref{sec:methodology}), except the ScanNet scenes taken directly from SceneSplat7K~\cite{li2025scenesplat}. For DL3DV-10K and ScanNet++, 49-frame sequences are rendered along smooth, interpolated camera trajectories, while ScanNet pairs ground-truth videos with noisy SceneSplat7K renders.
\vspace{-10pt}
\paragraph{Evaluation Protocol.} Following Difix3D+~\cite{Wu_2025_CVPR}, we evaluate trajectory-level video refinement on the DL3DV Benchmark~\cite{Ling_2024_CVPR} across three difficulty regimes defined by the number of training views and whether the test trajectory overlaps one of them: 
\textit{Low} (9 views, 1 overlapping), 
\textit{Medium} (3 views, 1 overlapping), and 
\textit{High} (6 views, no overlap with test trajectory) difficulties. 
Models refine 49-frame trajectories rendered at $832\times480$ resolution, which are subsequently used to reconstruct 3DGS representations. Appearance fidelity is evaluated on RGB renders via PSNR, SSIM, and LPIPS. Geometric accuracy is evaluated on depth renders via Absolute Relative Error ($\text{AbsRel}$), Root Mean Squared Error ($\text{RMSE}$), and threshold accuracies $\delta_k$ ($k \in \{1,2,3\}$) at $1.25^k$ tolerances~\cite{eigen2014depth}.
\vspace{-15pt}
\paragraph{Baselines.} We compare our model \textbf{Wan + Ctrl (Morph.)} against Difix and Fixer~\cite{Wu_2025_CVPR}, state-of-the-art standalone image-to-image refinement models trained to specifically fix 3DGS rendering artifacts. To evaluate these single-image baselines on trajectory-level sequences, we process every test frame independently. In addition, we compare against Wan+Tile \cite{TheDenk}, an off-the-shelf Wan2.2-based video refinement model sharing our ControlNet \cite{zhang2023adding} architecture trained on spatial-temporal tiling strategies.
\vspace{-10pt}
\paragraph{Results.} Table~\ref{tab:eccv_heatmap_views_subtle} and Figure~\ref{fig:qualitative_results} evaluate 3DGS reconstructions trained on refined trajectory videos across difficulty regimes. Comparing our \textbf{Wan + Ctrl (Morph.)} against \textbf{Wan-Tile} highlights the critical role of our training strategy: while \textbf{Wan-Tile} fails on both RGB appearance and depth geometry—filling artifacted regions with imagined details that significantly stray from the original scene—our morphologically perturbed dataset guides the ControlNet far more effectively. By conditioning on 3D morphological corruptions, \textbf{Wan + Ctrl (Morph.)} achieves superior artifact refinement and inpainting, generating details that blend naturally with the original scene while enforcing multi-view coherence. This translates into top performance across nearly all geometric metrics ($\text{AbsRel}$, $\text{RMSE}$, $\delta_{1,2,3}$) in \textit{Medium} and \textit{High} difficulty regimes, lowering depth RMSE to 16.18, 30.65, and 29.53, respectively. Single-image baselines further underscore this trade-off: \textit{Difix} produces sharp renders but fails to adequately refine structural artifacts, whereas \textit{Fixer} removes artifacts at the cost of severe blurriness. Crucially, because these single-frame baselines lack cross-frame temporal awareness, using their refined outputs for 3DGS reconstruction leads to geometric degradation, whereas \textbf{Wan + Ctrl (Morph.)} maintains visual fidelity while delivering structurally accurate 3D representations. Additional qualitative results across all difficulty settings are available in \textit{Supplementary Material}.

\begin{table}[t]
\centering
\caption{Success rate comparison across data augmentation methods for manipulation task episode generation (Mean$\uparrow \pm$ Std$\downarrow$ and Max$\uparrow$). \textbf{DREMA+ref.} uses refinement via \textbf{Wan+Ctrl(Morph.)}.}
\scriptsize
\setlength{\tabcolsep}{1.8pt}
\renewcommand{\arraystretch}{1.0}
\begin{tabular}{@{} l c c | c c | c c | c c @{}}
\toprule
\textbf{Method} & \multicolumn{2}{c|}{\textbf{Close Jar}} & \multicolumn{2}{c|}{\textbf{Insert Peg}} & \multicolumn{2}{c|}{\textbf{Lift}} & \multicolumn{2}{c}{\textbf{Pick Cup}} \\
& Mean$\pm$Std & Max & Mean$\pm$Std & Max & Mean$\pm$Std & Max & Mean$\pm$Std & Max \\
\midrule
PerAct & 38.4$\pm$0.80 & 40 & 0.0$\pm$0.00 & 0 & 22.8$\pm$1.60 & \textbf{26} & 13.2$\pm$2.04 & 16 \\
Patches & 45.2$\pm$3.49 & 48 & \cellcolor{green!10}\textit{2.0$\pm$1.79} & \cellcolor{green!10}\textit{4} & 20.4$\pm$2.65 & 24 & 37.2$\pm$3.25 & 42 \\
Table color & 45.0$\pm$1.73 & 46 & 1.6$\pm$1.50 & 4 & 23.6$\pm$0.80 & 24 & 25.6$\pm$2.94 & 30 \\
Distractors & 36.4$\pm$0.80 & 38 & 0.4$\pm$0.80 & 2 & 22.8$\pm$1.60 & 24 & \cellcolor{green!20}\textbf{41.2$\pm$2.99} & \cellcolor{green!20}\textbf{44} \\
DREMA & \cellcolor{green!10}\textit{51.2$\pm$1.60} & \cellcolor{green!10}\textit{54} & \cellcolor{green!20}\textbf{2.4$\pm$2.33} & \cellcolor{green!20}\textbf{6} & \cellcolor{green!10}\textit{23.6$\pm$1.50} & \cellcolor{green!10}\textit{\textbf{26}} & 34.4$\pm$3.88 & 40 \\
DREMA+ref. & \cellcolor{green!20}\textbf{59.2$\pm$5.45} & \cellcolor{green!20}\textbf{66} & 1.6$\pm$1.49 & 4 & \cellcolor{green!20}\textbf{24.4$\pm$1.96} & \cellcolor{green!20}\textbf{24} & \cellcolor{green!10}\textit{38.0$\pm$2.19} & \cellcolor{green!10}\textit{40} \\
\bottomrule
\end{tabular}
\label{tab:task_performance_compact}
\end{table}

\subsection{Downstream Robotics Imitation Learning}
\label{sec:robotics_downstream}

\paragraph{Setup.}
We evaluate our \textbf{Wan + Ctrl (Morph.)} model (Sec.~\ref{sec:3dgtr}) on downstream imitation learning for tabletop manipulation. As our baseline, we use Dream-to-Manipulate (DREMA)~\cite{barcellona2025dream}, which reconstructs scenes as Gaussians, segments objects, and recomposes them in PyBullet~\cite{coumans2016pybullet} to synthesize novel interaction episodes. While raw simulator trajectories provide precise geometry, they suffer from flat textures and synthetic lighting. We integrate our model into DREMA by refining simulator-rendered trajectories prior to Gaussian reconstruction. Rather than fixing structural errors, our goal is enhancing renders with realistic lighting and natural textures while maintaining strict 3D consistency for subsequent scenario generation. Following DREMA, we train PerAct~\cite{shridhar2022peract} agents on refined episodes across multi-view RLBench~\cite{james2019rlbench} camera streams (\textit{front}, \textit{left/right shoulder}, \textit{wrist}).

\vspace{-10pt}
\paragraph{Simulation Tasks.}
We evaluate policies across four RLBench tabletop manipulation tasks: \textit{Close Jar}, \textit{Insert Peg}, \textit{Lift}, and \textit{Pick Cup}. Crucially, we exclude tasks featuring fixed, static goal objects (e.g., \textit{Place Wine}, \textit{Sort Shape}), as zero appearance variation across episodes causes policies to overfit to static visual cues rather than spatial geometry.

\vspace{-10pt}
\paragraph{Baselines \& Metrics.}
We compare our pipeline (\textit{DREMA + ref.}) directly against standard \textit{DREMA}, unaugmented \textit{PerAct}, and classical augmentations reported in DREMA (\textit{Random Patches}~\cite{laskin2020reinforcement}, \textit{GenAug}~\cite{chen2023genaug}, and \textit{RoboAgent Distractors}~\cite{bharadhwaj2024roboagent}). Policies are trained for 100K iterations and evaluated over 50 test episodes across 5 seeds to report mean, standard deviation, and maximum success rates.

\vspace{-10pt}
\paragraph{Results.}
Table~\ref{tab:task_performance_compact} compares policy performance across four manipulation tasks. Integrating our video refinement into baseline DREMA improves downstream policy performance, yielding gains of up to 8.0\% over DREMA across 3 out of 4 tasks. Specifically, \textit{DREMA+ref.} achieves higher mean success rates than all evaluated baselines on \textit{Close Jar} (59.2\% vs. 51.2\% for DREMA) and \textit{Lift} (24.4\% vs. 23.6\%), while also improving over DREMA on \textit{Pick Cup} (38.0\% vs. 34.4\%). On precision-sensitive tasks like \textit{Insert Peg}, performance remains constrained across all methods due to physical execution constraints rather than visual fidelity. As shown in Figure~\ref{fig:qualitative_rlbench}, our refinement pipeline introduces photorealistic lighting dynamics, color shifts, and table textures to synthetic PyBullet renders. By providing appearance diversity while enforcing strict 3D geometric consistency, our approach helps prevent policies from overfitting to flat simulator artifacts and encourages PerAct to rely on underlying scene geometry.
\section{Conclusion}
\label{sec:conclusion}
Leveraging the explicit nature of 3D Gaussian Splatting, we propose \textbf{3D Morphological Perturbations}—an optimization-free augmentation that bypasses repeated per-scene reconstructions to obtain corrupted 3D scenes. By perturbing scale, rotation, and pruning, our approach acts as a regularizer for generative model training, outperforming 2D blur and alternative perturbation strategies. Scaled to a 14B-parameter video model via ControlNet, our method achieves a 12.5\% reduction in mean depth error over image-based refiners and translates these geometric gains into an 8.0\% boost in downstream robotics policy success. Our results highlight the effectiveness of 3D perturbations on increasing dataset curation efficiency for 3D-aware generative model training. We anticipate future work will extend these principles to other explicit 3D representations and implement them for on-the-fly training augmentations.

{
    \small
    \bibliographystyle{ieeenat_fullname}
    \bibliography{main}

@String(CVPR  = {IEEE Conf. Comput. Vis. Pattern Recog.})

@String(ICCV  = {Int. Conf. Comput. Vis.})

@String(ECCV  = {Eur. Conf. Comput. Vis.})

@String(CVPR  = {CVPR})

@String(ICCV  = {ICCV})

@String(ECCV  = {ECCV})

@String(CVPR= {IEEE Conf. Comput. Vis. Pattern Recog.})

@String(ICCV= {Int. Conf. Comput. Vis.})

@String(ECCV= {Eur. Conf. Comput. Vis.})

@inproceedings{mildenhall2020nerf,
 title={NeRF: Representing Scenes as Neural Radiance Fields for View Synthesis},
 author={Ben Mildenhall and Pratul P. Srinivasan and Matthew Tancik and Jonathan T. Barron and Ravi Ramamoorthi and Ren Ng},
 year={2020},
 booktitle={ECCV},
}

@Article{kerbl3Dgaussians,
      author       = {Kerbl, Bernhard and Kopanas, Georgios and Leimk{\"u}hler, Thomas and Drettakis, George},
      title        = {3D Gaussian Splatting for Real-Time Radiance Field Rendering},
      journal      = {ACM Transactions on Graphics},
      number       = {4},
      volume       = {42},
      month        = {July},
      year         = {2023},
      url          = {https://repo-sam.inria.fr/fungraph/3d-gaussian-splatting/}
}

@inproceedings{zhou2023nerflix,
title={NeRFLiX: High-Quality Neural View Synthesis by Learning a Degradation-Driven Inter-viewpoint MiXer},
author={Zhou, Kun and Li, Wenbo and Wang, Yi and Hu, Tao and Jiang, Nianjuan and Han, Xiaoguang and Lu, Jiangbo},
booktitle={Proceedings of the IEEE/CVF Conference on Computer Vision and Pattern Recognition},
pages={12363--12374},
year={2023}
}

@ARTICLE{10361604,
  author={Zhou, Kun and Li, Wenbo and Jiang, Nianjuan and Han, Xiaoguang and Lu, Jiangbo},
  journal={IEEE Transactions on Pattern Analysis and Machine Intelligence}, 
  title={From NeRFLiX to NeRFLiX++: A General NeRF-Agnostic Restorer Paradigm}, 
  year={2024},
  volume={46},
  number={5},
  pages={3422-3437},
  doi={10.1109/TPAMI.2023.3343395}}

@InProceedings{Wu_2025_CVPR,
    author    = {Wu, Jay Zhangjie and Zhang, Yuxuan and Turki, Haithem and Ren, Xuanchi and Gao, Jun and Shou, Mike Zheng and Fidler, Sanja and Gojcic, Zan and Ling, Huan},
    title     = {DIFIX3D+: Improving 3D Reconstructions with Single-Step Diffusion Models},
    booktitle = {Proceedings of the Computer Vision and Pattern Recognition Conference (CVPR)},
    month     = {June},
    year      = {2025},
    pages     = {26024-26035}
}

@inproceedings{NEURIPS2024_f0b42291,
 author = {Liu, Xi and Zhou, Chaoyi and Huang, Siyu},
 booktitle = {Advances in Neural Information Processing Systems},
 doi = {10.52202/079017-4237},
 editor = {A. Globerson and L. Mackey and D. Belgrave and A. Fan and U. Paquet and J. Tomczak and C. Zhang},
 pages = {133305--133327},
 publisher = {Curran Associates, Inc.},
 title = {3DGS-Enhancer: Enhancing Unbounded 3D Gaussian Splatting with View-consistent 2D Diffusion Priors},
 url = {https://proceedings.neurips.cc/paper_files/paper/2024/file/f0b42291ddab77dcb2ef8a3488301b62-Paper-Conference.pdf},
 volume = {37},
 year = {2024}
}

@article{Yu2024LMGaussianBS,
  title={LM-Gaussian: Boost Sparse-view 3D Gaussian Splatting with Large Model Priors},
  author={Hanyang Yu and Xiaoxiao Long and Ping Tan},
  journal={ArXiv},
  year={2024},
  volume={abs/2409.03456},
  url={https://api.semanticscholar.org/CorpusID:272423748}
}

@article{yin2025gsfixer,
  title={GSFixer: Improving 3D Gaussian Splatting with Reference-Guided Video Diffusion Priors},
  author={Yin, Xingyilang and Zhang, Qi and Chang, Jiahao and Feng, Ying and Fan, Qingnan and Yang, Xi and Pun, Chi-Man and Zhang, Huaqi and Cun, Xiaodong},
  journal={arXiv preprint arXiv:2508.09667},
  year={2025}
}

@inproceedings{wang2025vggt,
  title={VGGT: Visual Geometry Grounded Transformer},
  author={Wang, Jianyuan and Chen, Minghao and Karaev, Nikita and Vedaldi, Andrea and Rupprecht, Christian and Novotny, David},
  booktitle={Proceedings of the IEEE/CVF Conference on Computer Vision and Pattern Recognition},
  year={2025}
}

@article{
oquab2024dinov,
title={{DINO}v2: Learning Robust Visual Features without Supervision},
author={Maxime Oquab and Timoth{\'e}e Darcet and Th{\'e}o Moutakanni and Huy V. Vo and Marc Szafraniec and Vasil Khalidov and Pierre Fernandez and Daniel HAZIZA and Francisco Massa and Alaaeldin El-Nouby and Mido Assran and Nicolas Ballas and Wojciech Galuba and Russell Howes and Po-Yao Huang and Shang-Wen Li and Ishan Misra and Michael Rabbat and Vasu Sharma and Gabriel Synnaeve and Hu Xu and Herve Jegou and Julien Mairal and Patrick Labatut and Armand Joulin and Piotr Bojanowski},
journal={Transactions on Machine Learning Research},
issn={2835-8856},
year={2024},
url={https://openreview.net/forum?id=a68SUt6zFt},
note={Featured Certification}
}

@InProceedings{Ling_2024_CVPR,
    author    = {Ling, Lu and Sheng, Yichen and Tu, Zhi and Zhao, Wentian and Xin, Cheng and Wan, Kun and Yu, Lantao and Guo, Qianyu and Yu, Zixun and Lu, Yawen and Li, Xuanmao and Sun, Xingpeng and Ashok, Rohan and Mukherjee, Aniruddha and Kang, Hao and Kong, Xiangrui and Hua, Gang and Zhang, Tianyi and Benes, Bedrich and Bera, Aniket},
    title     = {DL3DV-10K: A Large-Scale Scene Dataset for Deep Learning-based 3D Vision},
    booktitle = {Proceedings of the IEEE/CVF Conference on Computer Vision and Pattern Recognition (CVPR)},
    month     = {June},
    year      = {2024},
    pages     = {22160-22169}
}

@article{wan2025wan,
  title={Wan: Open and advanced large-scale video generative models},
  author={Wan, Team and Wang, Ang and Ai, Baole and Wen, Bin and Mao, Chaojie and Xie, Chen-Wei and Chen, Di and Yu, Feiwu and Zhao, Haiming and Yang, Jianxiao and others},
  journal={arXiv preprint arXiv:2503.20314},
  year={2025}
}

@misc{zhang2023adding,
  title={Adding Conditional Control to Text-to-Image Diffusion Models}, 
  author={Lvmin Zhang and Anyi Rao and Maneesh Agrawala},
  booktitle={IEEE International Conference on Computer Vision (ICCV)},
  year={2023},
}

@inproceedings{dai2017scannet,
    title={ScanNet: Richly-annotated 3D Reconstructions of Indoor Scenes},
    author={Dai, Angela and Chang, Angel X. and Savva, Manolis and Halber, Maciej and Funkhouser, Thomas and Nie{\ss}ner, Matthias},
    booktitle = {Proc. Computer Vision and Pattern Recognition (CVPR), IEEE},
    year = {2017}
}

@inproceedings{yeshwanth2023scannet++,
  title={Scannet++: A high-fidelity dataset of 3d indoor scenes},
  author={Yeshwanth, Chandan and Liu, Yueh-Cheng and Nie{\ss}ner, Matthias and Dai, Angela},
  booktitle={Proceedings of the IEEE/CVF International Conference on Computer Vision},
  pages={12--22},
  year={2023}
}

@inproceedings{li2025scenesplat,
  title={SceneSplat: Gaussian Splatting-based Scene Understanding With Vision-Language Pretraining},
  author={Li, Yue and Ma, Qi and Yang, Runyi and Li, Huapeng and Ma, Mengjiao and Ren, Bin and Popovic, Nikola and Sebe, Nicu and Konukoglu, Ender and Gevers, Theo and others},
  booktitle = {Proceedings of the IEEE/CVF International Conference on Computer Vision (ICCV)},
  year={2025}
}

@misc{yang2024qwen2technicalreport,
      title={Qwen2 Technical Report}, 
      author={An Yang and Baosong Yang and Binyuan Hui and Bo Zheng and Bowen Yu and Chang Zhou and Chengpeng Li and Chengyuan Li and Dayiheng Liu and Fei Huang and Guanting Dong and Haoran Wei and Huan Lin and Jialong Tang and Jialin Wang and Jian Yang and Jianhong Tu and Jianwei Zhang and Jianxin Ma and Jianxin Yang and Jin Xu and Jingren Zhou and Jinze Bai and Jinzheng He and Junyang Lin and Kai Dang and Keming Lu and Keqin Chen and Kexin Yang and Mei Li and Mingfeng Xue and Na Ni and Pei Zhang and Peng Wang and Ru Peng and Rui Men and Ruize Gao and Runji Lin and Shijie Wang and Shuai Bai and Sinan Tan and Tianhang Zhu and Tianhao Li and Tianyu Liu and Wenbin Ge and Xiaodong Deng and Xiaohuan Zhou and Xingzhang Ren and Xinyu Zhang and Xipin Wei and Xuancheng Ren and Xuejing Liu and Yang Fan and Yang Yao and Yichang Zhang and Yu Wan and Yunfei Chu and Yuqiong Liu and Zeyu Cui and Zhenru Zhang and Zhifang Guo and Zhihao Fan},
      year={2024},
      eprint={2407.10671},
      archivePrefix={arXiv},
      primaryClass={cs.CL},
      url={https://arxiv.org/abs/2407.10671}, 
}

@inproceedings{
yang2025cogvideox,
title={CogVideoX: Text-to-Video Diffusion Models with An Expert Transformer},
author={Zhuoyi Yang and Jiayan Teng and Wendi Zheng and Ming Ding and Shiyu Huang and Jiazheng Xu and Yuanming Yang and Wenyi Hong and Xiaohan Zhang and Guanyu Feng and Da Yin and Yuxuan.Zhang and Weihan Wang and Yean Cheng and Bin Xu and Xiaotao Gu and Yuxiao Dong and Jie Tang},
booktitle={The Thirteenth International Conference on Learning Representations},
year={2025},
url={https://openreview.net/forum?id=LQzN6TRFg9}
}

@article{gsfix3d,
         title={GSFix3D: Diffusion-Guided Repair of Novel Views in Gaussian Splatting}, 
         author={Jiaxin Wei and Stefan Leutenegger and Simon Schaefer},
         year={2025},
         eprint={2508.14717},
         archivePrefix={arXiv},
         primaryClass={cs.CV},
         url={https://arxiv.org/abs/2508.14717},
}

@InProceedings{10.1007/978-3-031-72640-8_19,
author="Liu, Xinhang
and Chen, Jiaben
and Kao, Shiu-Hong
and Tai, Yu-Wing
and Tang, Chi-Keung",
editor="Leonardis, Ale{\v{s}}
and Ricci, Elisa
and Roth, Stefan
and Russakovsky, Olga
and Sattler, Torsten
and Varol, G{\"u}l",
title="Deceptive-NeRF/3DGS: Diffusion-Generated Pseudo-observations for High-Quality Sparse-View Reconstruction",
booktitle="Computer Vision -- ECCV 2024",
year="2025",
publisher="Springer Nature Switzerland",
address="Cham",
pages="337--355",
isbn="978-3-031-72640-8"
}

@InProceedings{Paliwal_2025_ICCV,
    author    = {Paliwal, Avinash and Zhou, Xilong and Ye, Wei and Xiong, Jinhui and Ranjan, Rakesh and Kalantari, Nima Khademi},
    title     = {RI3D: Few-Shot Gaussian Splatting With Repair and Inpainting Diffusion Priors},
    booktitle = {Proceedings of the IEEE/CVF International Conference on Computer Vision (ICCV)},
    month     = {October},
    year      = {2025},
    pages     = {25094-25103}
}

@InProceedings{Zhong_2025_CVPR,
    author    = {Zhong, Yingji and Li, Zhihao and Chen, Dave Zhenyu and Hong, Lanqing and Xu, Dan},
    title     = {Taming Video Diffusion Prior with Scene-Grounding Guidance for 3D Gaussian Splatting from Sparse Inputs},
    booktitle = {Proceedings of the IEEE/CVF Conference on Computer Vision and Pattern Recognition (CVPR)},
    month     = {June},
    year      = {2025},
    pages     = {6133-6143}
}

@misc{yu2025lmgaussianboostsparseview3d,
      title={LM-Gaussian: Boost Sparse-view 3D Gaussian Splatting with Large Model Priors}, 
      author={Hanyang Yu and Xiaoxiao Long and Ping Tan},
      year={2025},
      eprint={2409.03456},
      archivePrefix={arXiv},
      primaryClass={cs.CV},
      url={https://arxiv.org/abs/2409.03456}, 
}

@inproceedings{
barcellona2025dream,
title={Dream to Manipulate: Compositional World Models Empowering Robot Imitation Learning with Imagination},
author={Leonardo Barcellona and Andrii Zadaianchuk and Davide Allegro and Samuele Papa and Stefano Ghidoni and Efstratios Gavves},
booktitle={The Thirteenth International Conference on Learning Representations},
year={2025},
url={https://openreview.net/forum?id=3RSLW9YSgk}
}

@article{james2019rlbench,
  title={RLBench: The Robot Learning Benchmark \& Learning Environment},
  author={James, Stephen and Ma, Zicong and Rovick Arrojo, David and Davison, Andrew J.},
  journal={IEEE Robotics and Automation Letters},
  year={2020}
}

@InProceedings{rombach2021highresolution,
    author    = {Rombach, Robin and Blattmann, Andreas and Lorenz, Dominik and Esser, Patrick and Ommer, Bj\"orn},
    title     = {High-Resolution Image Synthesis With Latent Diffusion Models},
    booktitle = {Proceedings of the IEEE/CVF Conference on Computer Vision and Pattern Recognition (CVPR)},
    month     = {June},
    year      = {2022},
    pages     = {10684-10695}
}

@inproceedings{
hu2022lora,
title={Lo{RA}: Low-Rank Adaptation of Large Language Models},
author={Edward J Hu and Yelong Shen and Phillip Wallis and Zeyuan Allen-Zhu and Yuanzhi Li and Shean Wang and Lu Wang and Weizhu Chen},
booktitle={International Conference on Learning Representations},
year={2022},
url={https://openreview.net/forum?id=nZeVKeeFYf9}
}

@article{eigen2014depth,
  title={Depth map prediction from a single image using a multi-scale deep network},
  author={Eigen, David and Puhrsch, Christian and Fergus, Rob},
  journal={Advances in neural information processing systems},
  volume={27},
  year={2014}
}

@misc{coumans2016pybullet,
  title={Pybullet, a python module for physics simulation for games, robotics and machine learning},
  author={Coumans, Erwin and Bai, Yunfei},
  year={2016}
}

@article{laskin2020reinforcement,
  title={Reinforcement learning with augmented data},
  author={Laskin, Misha and Lee, Kimin and Stooke, Adam and Pinto, Lerrel and Abbeel, Pieter and Srinivas, Aravind},
  journal={Advances in neural information processing systems},
  volume={33},
  pages={19884--19895},
  year={2020}
}

@article{chen2023genaug,
  title={Genaug: Retargeting behaviors to unseen situations via generative augmentation},
  author={Chen, Zoey and Kiami, Sho and Gupta, Abhishek and Kumar, Vikash},
  journal={arXiv preprint arXiv:2302.06671},
  year={2023}
}

@inproceedings{bharadhwaj2024roboagent,
  title={Roboagent: Generalization and efficiency in robot manipulation via semantic augmentations and action chunking},
  author={Bharadhwaj, Homanga and Vakil, Jay and Sharma, Mohit and Gupta, Abhinav and Tulsiani, Shubham and Kumar, Vikash},
  booktitle={2024 IEEE International Conference on Robotics and Automation (ICRA)},
  pages={4788--4795},
  year={2024},
  organization={IEEE}
}

@inproceedings{
guo2024animatediff,
title={AnimateDiff: Animate Your Personalized Text-to-Image Diffusion Models without Specific Tuning},
author={Yuwei Guo and Ceyuan Yang and Anyi Rao and Zhengyang Liang and Yaohui Wang and Yu Qiao and Maneesh Agrawala and Dahua Lin and Bo Dai},
booktitle={The Twelfth International Conference on Learning Representations},
year={2024},
url={https://openreview.net/forum?id=Fx2SbBgcte}
}

@InProceedings{pmlr-v97-kornblith19a,
  title = 	 {Similarity of Neural Network Representations Revisited},
  author =       {Kornblith, Simon and Norouzi, Mohammad and Lee, Honglak and Hinton, Geoffrey},
  booktitle = 	 {Proceedings of the 36th International Conference on Machine Learning},
  pages = 	 {3519--3529},
  year = 	 {2019},
  editor = 	 {Chaudhuri, Kamalika and Salakhutdinov, Ruslan},
  volume = 	 {97},
  series = 	 {Proceedings of Machine Learning Research},
  month = 	 {09--15 Jun},
  publisher =    {PMLR},
  url = 	 {https://proceedings.mlr.press/v97/kornblith19a.html}
}

@ARTICLE{fovnerf,
  author={Deng, Nianchen and He, Zhenyi and Ye, Jiannan and Duinkharjav, Budmonde and Chakravarthula, Praneeth and Yang, Xubo and Sun, Qi},
  journal={IEEE Transactions on Visualization and Computer Graphics}, 
  title={FoV-NeRF: Foveated Neural Radiance Fields for Virtual Reality}, 
  year={2022},
  volume={28},
  number={11},
  pages={3854-3864},
  doi={10.1109/TVCG.2022.3203102}}

@inproceedings{10.1145/3641519.3657448,
author = {Jiang, Ying and Yu, Chang and Xie, Tianyi and Li, Xuan and Feng, Yutao and Wang, Huamin and Li, Minchen and Lau, Henry and Gao, Feng and Yang, Yin and Jiang, Chenfanfu},
title = {VR-GS: A Physical Dynamics-Aware Interactive Gaussian Splatting System in Virtual Reality},
year = {2024},
isbn = {9798400705250},
publisher = {Association for Computing Machinery},
address = {New York, NY, USA},
url = {https://doi.org/10.1145/3641519.3657448},
doi = {10.1145/3641519.3657448},
booktitle = {ACM SIGGRAPH 2024 Conference Papers},
articleno = {78},
numpages = {1},
location = {Denver, CO, USA},
series = {SIGGRAPH '24}
}

@INPROCEEDINGS{11092736,
  author={Chen, Jianchuan and Hu, Jingchuan and Wang, Gaige and Jiang, Zhonghua and Zhou, Tiansong and Chen, Zhiwen and Lv, Chengfei},
  booktitle={2025 IEEE/CVF Conference on Computer Vision and Pattern Recognition (CVPR)}, 
  title={TaoAvatar: Real-Time Lifelike Full-Body Talking Avatars for Augmented Reality via 3D Gaussian Splatting}, 
  year={2025},
  volume={},
  number={},
  pages={10723-10734},
  doi={10.1109/CVPR52734.2025.01002}}

@ARTICLE{9712211,
  author={Adamkiewicz, Michal and Chen, Timothy and Caccavale, Adam and Gardner, Rachel and Culbertson, Preston and Bohg, Jeannette and Schwager, Mac},
  journal={IEEE Robotics and Automation Letters}, 
  title={Vision-Only Robot Navigation in a Neural Radiance World}, 
  year={2022},
  volume={7},
  number={2},
  pages={4606-4613},
  doi={10.1109/LRA.2022.3150497}}

@article{10.1109/TRO.2025.3552348,
author = {Chen, Timothy and Shorinwa, Ola and Bruno, Joseph and Swann, Aiden and Yu, Javier and Zeng, Weijia and Nagami, Keiko and Dames, Philip and Schwager, Mac},
title = {Splat-Nav: Safe Real-Time Robot Navigation in Gaussian Splatting Maps},
year = {2025},
issue_date = {2025},
publisher = {IEEE Press},
volume = {41},
issn = {1552-3098},
url = {https://doi.org/10.1109/TRO.2025.3552348},
doi = {10.1109/TRO.2025.3552348},
journal = {Trans. Rob.},
month = jan,
pages = {2765–2784},
numpages = {20}
}

@INPROCEEDINGS{10160842,
  author={Dai, Qiyu and Zhu, Yan and Geng, Yiran and Ruan, Ciyu and Zhang, Jiazhao and Wang, He},
  booktitle={2023 IEEE International Conference on Robotics and Automation (ICRA)}, 
  title={GraspNeRF: Multiview-based 6-DoF Grasp Detection for Transparent and Specular Objects Using Generalizable NeRF}, 
  year={2023},
  volume={},
  number={},
  pages={1757-1763},
  doi={10.1109/ICRA48891.2023.10160842}}

@article{chen2026survey,
  title={A survey on 3d gaussian splatting},
  author={Chen, Guikun and Wang, Wenguan},
  journal={ACM Computing Surveys},
  volume={58},
  number={12},
  pages={1--39},
  year={2026},
  publisher={ACM New York, NY}
}

@article{he2026survey,
  title={A survey on 3d gaussian splatting applications: Segmentation, editing, and generation},
  author={He, Shuting and Ji, Peilin and Yang, Yitong and Wang, Changshuo and Ji, Jiayi and Wang, Yinglin and Ding, Henghui},
  journal={IEEE Transactions on Pattern Analysis and Machine Intelligence},
  year={2026},
  publisher={IEEE}
}

@inproceedings{li2024dngaussian,
  title={Dngaussian: Optimizing sparse-view 3d gaussian radiance fields with global-local depth normalization},
  author={Li, Jiahe and Zhang, Jiawei and Bai, Xiao and Zheng, Jin and Ning, Xin and Zhou, Jun and Gu, Lin},
  booktitle={2024 IEEE/CVF Conference on Computer Vision and Pattern Recognition (CVPR)},
  pages={20775--20785},
  year={2024},
  organization={IEEE}
}

@inproceedings{zhu2024fsgs,
  title={Fsgs: Real-time few-shot view synthesis using gaussian splatting},
  author={Zhu, Zehao and Fan, Zhiwen and Jiang, Yifan and Wang, Zhangyang},
  booktitle={European conference on computer vision},
  pages={145--163},
  year={2024},
  organization={Springer}
}

@inproceedings{chen2024mvsplat,
  title={Mvsplat: Efficient 3d gaussian splatting from sparse multi-view images},
  author={Chen, Yuedong and Xu, Haofei and Zheng, Chuanxia and Zhuang, Bohan and Pollefeys, Marc and Geiger, Andreas and Cham, Tat-Jen and Cai, Jianfei},
  booktitle={European conference on computer vision},
  pages={370--386},
  year={2024},
  organization={Springer}
}

@inproceedings{charatan2024pixelsplat,
  title={pixelsplat: 3d gaussian splats from image pairs for scalable generalizable 3d reconstruction},
  author={Charatan, David and Li, Sizhe Lester and Tagliasacchi, Andrea and Sitzmann, Vincent},
  booktitle={2024 IEEE/CVF Conference on Computer Vision and Pattern Recognition (CVPR)},
  pages={19457--19467},
  year={2024},
  organization={IEEE}
}

@inproceedings{szymanowicz2024splatter,
  title={Splatter image: Ultra-fast single-view 3d reconstruction},
  author={Szymanowicz, Stanislaw and Rupprecht, Chrisitian and Vedaldi, Andrea},
  booktitle={Proceedings of the IEEE/CVF conference on computer vision and pattern recognition},
  pages={10208--10217},
  year={2024}
}

@inproceedings{szymanowicz2025flash3d,
  title={Flash3d: Feed-forward generalisable 3d scene reconstruction from a single image},
  author={Szymanowicz, Stanislaw and Insafutdinov, Eldar and Zheng, Chuanxia and Campbell, Dylan and Henriques, Joao F and Rupprecht, Christian and Vedaldi, Andrea},
  booktitle={2025 International Conference on 3D Vision (3DV)},
  pages={670--681},
  year={2025},
  organization={IEEE}
}

@inproceedings{xu2024grm,
  title={Grm: Large gaussian reconstruction model for efficient 3d reconstruction and generation},
  author={Xu, Yinghao and Shi, Zifan and Yifan, Wang and Chen, Hansheng and Yang, Ceyuan and Peng, Sida and Shen, Yujun and Wetzstein, Gordon},
  booktitle={European Conference on Computer Vision},
  pages={1--20},
  year={2024},
  organization={Springer}
}

@inproceedings{wu2025difix3d+,
  title={Difix3d+: Improving 3d reconstructions with single-step diffusion models},
  author={Wu, Jay Zhangjie and Zhang, Yuxuan and Turki, Haithem and Ren, Xuanchi and Gao, Jun and Shou, Mike Zheng and Fidler, Sanja and Gojcic, Zan and Ling, Huan},
  booktitle={2025 IEEE/CVF Conference on Computer Vision and Pattern Recognition (CVPR)},
  pages={26024--26035},
  year={2025},
  organization={IEEE}
}

@inproceedings{wu2025genfusion,
  title={Genfusion: Closing the loop between reconstruction and generation via videos},
  author={Wu, Sibo and Xu, Congrong and Huang, Binbin and Geiger, Andreas and Chen, Anpei},
  booktitle={2025 IEEE/CVF Conference on Computer Vision and Pattern Recognition (CVPR)},
  pages={6078--6088},
  year={2025},
  organization={IEEE}
}

@inproceedings{de2026artifixer,
  title={ArtiFixer: Enhancing and Extending 3D Reconstruction with Auto-Regressive Diffusion Models},
  author={De Lutio, Riccardo and Fischer, Tobias and Chang, Yen-Yu and Zhang, Yuxuan and Wu, Zhangjie and Ren, Xuanchi and Shen, Tianchang and T{\'o}thov{\'a}, Katar{\'\i}na and Gojcic, Zan and Turki, Haithem},
  booktitle={Proceedings of the Special Interest Group on Computer Graphics and Interactive Techniques Conference Conference Papers},
  pages={1--12},
  year={2026}
}

@ARTICLE{11176446,
author={Yu, Wangbo and Xing, Jinbo and Yuan, Li and Hu, Wenbo and Li, Xiaoyu and Huang, Zhipeng and Gao, Xiangjun and Wong, Tien-Tsin and Shan, Ying and Tian, Yonghong},
journal={ IEEE Transactions on Pattern Analysis \& Machine Intelligence },
title={{ ViewCrafter: Taming Video Diffusion Models for High-fidelity Novel View Synthesis }},
year={5555},
volume={},
number={01},
ISSN={1939-3539},
pages={1-18},
doi={10.1109/TPAMI.2025.3613256},
url = {https://doi.ieeecomputersociety.org/10.1109/TPAMI.2025.3613256},
publisher={IEEE Computer Society},
address={Los Alamitos, CA, USA},
month=sep}

@inproceedings{fischer2025flowr,
  title={Flowr: Flowing from sparse to dense 3d reconstructions},
  author={Fischer, Tobias and Bul{\`o}, Samuel Rota and Yang, Yung-Hsu and Keetha, Nikhil and Porzi, Lorenzo and M{\"u}ller, Norman and Schwarz, Katja and Luiten, Jonathon and Pollefeys, Marc and Kontschieder, Peter},
  booktitle={2025 IEEE/CVF International Conference on Computer Vision (ICCV)},
  pages={27702--27712},
  year={2025},
  organization={IEEE}
}

@article{shorten2019survey,
  title={A survey on image data augmentation for deep learning},
  author={Shorten, Connor and Khoshgoftaar, Taghi M},
  journal={Journal of big data},
  volume={6},
  number={1},
  pages={60},
  year={2019},
  publisher={Springer}
}

@article{cubuk2020randaugment,
  title={Randaugment: Practical automated data augmentation with a reduced search space},
  author={Cubuk, Ekin Dogus and Zoph, Barret and Shlens, Jon and Le, Quoc},
  journal={Advances in neural information processing systems},
  volume={33},
  pages={18613--18624},
  year={2020}
}

@inproceedings{ICLR2024_94894cf9,
 author = {GUO, Yuwei and Yang, Ceyuan and Rao, Anyi and Liang, Zhengyang and Wang, Yaohui and Qiao, Yu and Agrawala, Maneesh and Lin, Dahua and DAI, Bo},
 booktitle = {International Conference on Learning Representations},
 editor = {B. Kim and Y. Yue and S. Chaudhuri and K. Fragkiadaki and M. Khan and Y. Sun},
 pages = {34630--34648},
 title = {AnimateDiff: Animate Your Personalized Text-to-Image Diffusion Models without Specific Tuning},
 url = {https://proceedings.iclr.cc/paper_files/paper/2024/file/94894cf9bab20c00f6eaaff3b7f2be70-Paper-Conference.pdf},
 volume = {2024},
 year = {2024}
}

@ARTICLE{bishop1995training,
  author={Bishop, Chris M.},
  journal={Neural Computation}, 
  title={Training with Noise is Equivalent to Tikhonov Regularization}, 
  year={1995},
  volume={7},
  number={1},
  pages={108-116},
  doi={10.1162/neco.1995.7.1.108}}

@InProceedings{maaten2013learning,
  title = 	 {Learning with Marginalized Corrupted Features},
  author = 	 {Maaten, Laurens and Chen, Minmin and Tyree, Stephen and Weinberger, Kilian},
  booktitle = 	 {Proceedings of the 30th International Conference on Machine Learning},
  pages = 	 {410--418},
  year = 	 {2013},
  editor = 	 {Dasgupta, Sanjoy and McAllester, David},
  volume = 	 {28},
  number =       {1},
  series = 	 {Proceedings of Machine Learning Research},
  address = 	 {Atlanta, Georgia, USA},
  month = 	 {17--19 Jun},
  publisher =    {PMLR},
  url = 	 {https://proceedings.mlr.press/v28/vandermaaten13.html}
}

@inproceedings{wager2013dropout,
 author = {Wager, Stefan and Wang, Sida and Liang, Percy S},
 booktitle = {Advances in Neural Information Processing Systems},
 editor = {C.J. Burges and L. Bottou and M. Welling and Z. Ghahramani and K. Weinberger},
 pages = {},
 publisher = {Curran Associates, Inc.},
 title = {Dropout Training as Adaptive Regularization},
 url = {https://proceedings.neurips.cc/paper_files/paper/2013/file/38db3aed920cf82ab059bfccbd02be6a-Paper.pdf},
 volume = {26},
 year = {2013}
}

@misc{TheDenk,
    title={Wan2.2 Controlnet},
    author={Karachev Denis},
    url={https://github.com/TheDenk/wan2.2-controlnet},
    publisher={Github},
    year={2025}
}

@inproceedings{unified,
 author = {Xin, Yuelin and Liu, Yuheng and Xie, Xiaohui and Li, Xinke},
 booktitle = {International Conference on Learning Representations},
 editor = {C. Vondrick and B. Hariharan and C. Raffel and L. Pinto and D. Yang and A. Faust},
 pages = {1111--1123},
 title = {Learning Unified Representation of 3D Gaussian Splatting},
 url = {https://proceedings.iclr.cc/paper_files/paper/2026/file/02c080341e534c5a02566c8af0bf7137-Paper-Conference.pdf},
 volume = {2026},
 year = {2026}
}

@InProceedings{augnerf,
    author    = {Chen, Tianlong and Wang, Peihao and Fan, Zhiwen and Wang, Zhangyang},
    title     = {Aug-NeRF: Training Stronger Neural Radiance Fields With Triple-Level Physically-Grounded Augmentations},
    booktitle = {Proceedings of the IEEE/CVF Conference on Computer Vision and Pattern Recognition (CVPR)},
    month     = {June},
    year      = {2022},
    pages     = {15191-15202}
}

@inproceedings{hicom,
 author = {Gao, Qiankun and Meng, Jiarui and Wen, Chengxiang and Chen, Jie and Zhang, Jian},
 booktitle = {Advances in Neural Information Processing Systems},
 doi = {10.52202/079017-2563},
 editor = {A. Globerson and L. Mackey and D. Belgrave and A. Fan and U. Paquet and J. Tomczak and C. Zhang},
 pages = {80609--80633},
 publisher = {Curran Associates, Inc.},
 title = {HiCoM: Hierarchical Coherent Motion for Dynamic Streamable Scenes with 3D Gaussian Splatting},
 url = {https://proceedings.neurips.cc/paper_files/paper/2024/file/9370cc8d438b9ed8cb4344818a251d9a-Paper-Conference.pdf},
 volume = {37},
 year = {2024}
}

@inproceedings{recongs,
 author = {Fu, Jiaye and Gao, Qiankun and Wen, Chengxiang and Wu, Yanmin and Ma, Siwei and Zhang, Jiaqi and Zhang, Jian},
 booktitle = {Advances in Neural Information Processing Systems},
 doi = {10.52202/085713-4496},
 editor = {D. Belgrave and C. Zhang and H. Lin and R. Pascanu and P. Koniusz and M. Ghassemi and N. Chen},
 pages = {134751--134778},
 publisher = {Curran Associates, Inc.},
 title = {ReCon-GS: Continuum-Preserved Gaussian Streaming for Fast and Compact Reconstruction of Dynamic Scenes},
 url = {https://proceedings.neurips.cc/paper_files/paper/2025/file/c3e969ea20542a6a11e6caeac736a0b9-Paper-Conference.pdf},
 volume = {38, Main Conference},
 year = {2025}
}

@inproceedings{peebles2023scalable,
  title={Scalable diffusion models with transformers},
  author={Peebles, William and Xie, Saining},
  booktitle={2023 IEEE/CVF International Conference on Computer Vision (ICCV)},
  pages={4172--4182},
  year={2023},
  organization={IEEE}
}

@inproceedings{shridhar2022peract,
  title     = {Perceiver-Actor: A Multi-Task Transformer for Robotic Manipulation},
  author    = {Shridhar, Mohit and Manuelli, Lucas and Fox, Dieter},
  booktitle = {Proceedings of the 6th Conference on Robot Learning (CoRL)},
  year      = {2022},
}
}

% WARNING: do not forget to delete the supplementary pages from your submission 
\clearpage
\appendix
\renewcommand{\thefigure}{A\arabic{figure}}
\renewcommand{\thetable}{A\arabic{table}}
\setcounter{page}{1}
\setcounter{section}{0}
\setcounter{figure}{0}
\setcounter{table}{0}
\maketitlesupplementary

\section{Diagnostic Ablation: Perturbation Categories}
\label{sec:animatediff_supp}

This section complements Section \ref{sec:ablation} of the main paper. Section \ref{sec:animatediff_implementation} details the architectural modifications, optimization parameters, and evaluation protocols for our AnimateDiff-based lightweight video diffusion model \cite{guo2024animatediff}. Section \ref{sec:animatediff_supp_qualitative} presents extended qualitative comparisons within all sparsity regimes ($\text{S}_{20}$, $\text{S}_{40}$, and $\text{S}_{80}$).

\subsection{AnimateDiff + Stable Diffusion + LoRA Implementation Details for Video-to-Video Refinement}
\label{sec:animatediff_implementation}

We adopt Stable Diffusion v1.5~\cite{rombach2021highresolution} combined with the AnimateDiff motion adapter v1.5.2~\cite{guo2024animatediff} as our base 3D UNet backbone ($\mathcal{U}$). Standard diffusion UNets accept a 4-channel latent representation $z_t \in \mathbb{R}^{B \times C \times F \times H \times W}$ (where $C=4$). To inject spatial-temporal trajectory conditioning into a base UNet with no inherent 3D knowledge, we expand the input convolutional layer (\texttt{conv\_in}) from 4 to 8 channels to receive $z_{\text{cond}}$, directly anchoring all scene geometry and spatial alignment to our 3D Gaussian renderings.

The extended 8-channel tensor represents the channel-wise concatenation $[z_t, z_{\text{cond}}]$, where $z_{\text{cond}}$ contains the cached latents of noisy/perturbed 3D Gaussian renderings. The modified \texttt{conv\_in} layer weights $W_{\text{new}} \in \mathbb{R}^{C_{\text{out}} \times 8 \times 3 \times 3}$ are initialized via weight surgery on original weights $W_{\text{base}} \in \mathbb{R}^{C_{\text{out}} \times 4 \times 3 \times 3}$:

\begin{equation}
W_{\text{new}}[:, :4, \cdot, \cdot] = W_{\text{base}}, \quad W_{\text{new}}[:, 4:, \cdot, \cdot] = 0.1 \times W_{\text{base}}.
\end{equation}

Scaling the newly added conditioning channels by a factor of $0.1$ prevents initial magnitude explosion in intermediate UNet activations while enabling immediate gradient flow into $z_{\text{cond}}$ features from the first optimization step.

To retain base feature representations and minimize computational overhead, the 2D UNet backbone and 1D temporal motion modules remain frozen during training. Parameter updates are strictly restricted to Low-Rank Adaptation (LoRA)~\cite{hu2022lora} matrices injected into spatial and temporal attention blocks. Specifically, rank $r=64$ and scaling factor $\alpha=128$ (with dropout rate $p=0.05$) are applied to projection matrices across both spatial attention (\texttt{to\_q}, \texttt{to\_k}, \texttt{to\_v}, \texttt{to\_out.0}) and 1D temporal motion attention blocks (\texttt{motion\_attn.to\_q}, \texttt{motion\_attn.to\_k}, \texttt{motion\_attn.to\_v}, \texttt{motion\_attn.to\_out.0}), forcing the network to leverage $z_{\text{cond}}$ for spatial alignment while adapting only motion dynamics and surface appearance. Additionally, the non-pretrained 8-channel \texttt{conv\_in} layer parameters are explicitly set to \texttt{requires\_grad=True} and optimized concurrently with an elevated learning rate.

The complete training hyperparameter specification is detailed in Table~\ref{tab:animatediff_hyperparams}. Optimization is performed using AdamW with cosine learning rate scheduling and a $5\%$ linear warmup phase under Automatic Mixed Precision (\texttt{bfloat16}). 

To evaluate performance across varying camera displacement speeds, held-out validation is conducted across three temporal stride regimes: Stride 20 ($\text{S}_{20}$), Stride 40 ($\text{S}_{40}$), and Stride 80 ($\text{S}_{80}$).

\begin{table*}[h!]
\centering
\small
\setlength{\tabcolsep}{4pt}
\caption{Hyperparameter specification for AnimateDiff + Stable Diffusion LoRA fine-tuning.}
\label{tab:animatediff_hyperparams}
\begin{tabular}{l l}
\toprule
\textbf{Parameter} & \textbf{Value} \\
\midrule
Base Architecture & SD v1.5 + AnimateDiff v1.5.2 \\
Input Channels ($C_{\text{in}}$) & $8$ ($4\times\text{Noisy Latents} + 4\times\text{Conditioning}$) \\
LoRA Rank ($r$) / Alpha ($\alpha$) & $64$ / $128$ \\
LoRA Target Modules & Spatial \& Motion Attn. ($Q,K,V,\text{Out}$) \\
LoRA Dropout & $0.05$ \\
Learning Rate (LoRA / \texttt{conv\_in}) & $2 \times 10^{-4}$ / $2 \times 10^{-4}$ \\
Optimizer & AdamW ($\beta_1{=}0.9, \beta_2{=}0.999$, Weight Decay ${=} 10^{-2}$) \\
LR Schedule & Cosine w/ Warmup ($5\%$ total steps) \\
Batch Size / Grad. Accum. & $1$ / $4$ (Effective Batch Size $= 4$) \\
Precision & Automatic Mixed Precision (\texttt{bfloat16}) \\
Grad. Clipping Norm & $1.0$ \\
Training Epochs & $200$ \\
\bottomrule
\end{tabular}
\end{table*}

\subsection{Extended Qualitative Results}
\label{sec:animatediff_supp_qualitative}
We present extended qualitative comparisons for the ablation study detailed in Section~\ref{sec:ablation} of the main paper. Figure~\ref{fig:supp_s80} expands upon Figure~\ref{fig:animatediff_quantitative} from the main paper for the extreme sparsity setting ($\mathcal{S}_{80}$). Additionally, Figures~\ref{fig:supp_s40} and~\ref{fig:supp_s20} illustrate qualitative outputs under high ($\mathcal{S}_{40}$) and moderate ($\mathcal{S}_{20}$) input sparsity, respectively.

These expanded results align with our findings in Section~\ref{sec:ablation}, demonstrating consistent performance trends across all sparsity regimes. While higher input density ($\mathcal{S}_{40}$ and $\mathcal{S}_{20}$) generally improves quality across all models compared to $\mathcal{S}_{80}$, the model trained on \textbf{3D Morphological Perturbations} remains the most consistent across all sparsities. Notably, models trained on \textit{2D Render Blur} and \textit{3D Naive Perturbations} show the least improvement, further highlighting the critical role of perturbation selection.

\section{Scaling to Video Foundation Models for Trajectory Render Refinement}
\label{sec:wan_results}

This section complements Section~\ref{sec:3dgtr} of the main paper by providing extended implementation details and qualitative evaluations for our trajectory refinement framework. Specifically, Section~\ref{sec:wan_model_details} details the architecture, training formulation, and inference mechanics of our \textbf{Wan + ControlNet (Morph.)} model, while Section~\ref{sec:wan_additional_results} presents extended visual comparisons and depth renders across sparse-view difficulty regimes on the DL3DV-10K benchmark~\cite{Ling_2024_CVPR}.

\subsection{Wan + ControlNet (Morph.) Model Details}
\label{sec:wan_model_details}

In this section, we provide additional details on the implementation of our video-to-video diffusion refinement model, \textbf{Wan + ControlNet (Morph.)}, as well as its training and inference pipelines.

For the base model, we adopt Wan2.2's T2V-A14B model~\cite{wan2025wan}, a 14B-parameter text-to-video diffusion model capable of generating videos at 480P and 720P resolutions. In all experiments, both training and testing videos are formatted to 49-frame sequences resized to a resolution of $832 \times 480$.

\begin{figure}[t!]
      \centering
      \includegraphics[scale=0.43]{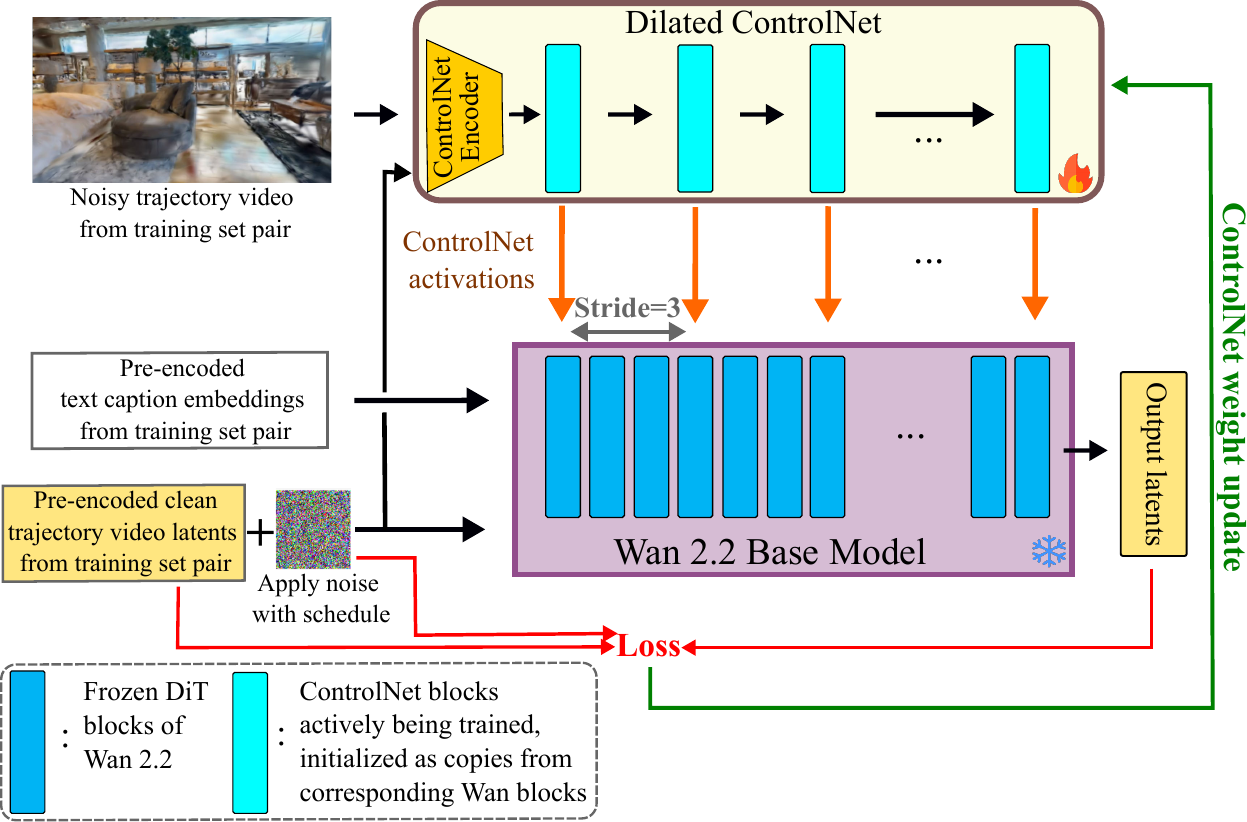}
      \caption{The architecture and training procedure of our \textbf{Wan + ControlNet (Morph.)} video-to-video diffusion refinement model.}
      \label{fig:supp_arch_train}
\end{figure}

The architecture and training pipeline are illustrated in Figure~\ref{fig:supp_arch_train}. Prior to training, text caption embeddings generated via Qwen2~\cite{yang2024qwen2technicalreport} and latent representations of the ground-truth trajectory videos are precomputed using Wan2.2's VAE encoder and cached to accelerate training.

As described in Section~\ref{sec:3dgtr} of the main paper, a dilated ControlNet~\cite{zhang2023adding} is initialized using eight DiT blocks copied from the base model with a dilation stride of 3. During training, random noise is added to the ground-truth latent representation $z_0$ using a Flow-Match Euler discrete scheduler, originally introduced in Stable Diffusion 3 [Esser \textit{et al.}, 2024]. Specifically, a timestep $t$ is sampled and Gaussian noise $\epsilon \sim \mathcal{N}(0, I)$ is added to obtain the noisy latent $z_t$.
%)~\cite{esser2024scalingrectifiedflowtransformers}

The noisy latent $z_t$ is provided as input to both the frozen base model and the ControlNet. In addition, the ControlNet receives the corresponding noisy trajectory video as conditioning input, while the base model is conditioned on the text caption embeddings. The ControlNet predicts activation offsets that are injected additively into the corresponding frozen blocks of the base model.

Conditioned on these activations, the base model predicts the noise residual $\hat{\epsilon}_\theta(z_t, t)$ required to reconstruct the clean latent. The model is trained using the standard noise prediction objective:

\begin{equation}
\mathcal{L} =
\mathbb{E}_{z_0, \epsilon, t}
\left[
\left\|
\hat{\epsilon}_\theta(z_t, t) - \epsilon
\right\|_2^2
\right].
\end{equation}

Gradients from this loss are backpropagated exclusively through the ControlNet parameters, while the base Wan2.2 model remains frozen. Training is conducted for 30 epochs across 8 GPUs with 80GB VRAM.

\begin{figure}[t!]
      \centering
      \includegraphics[scale=0.43]{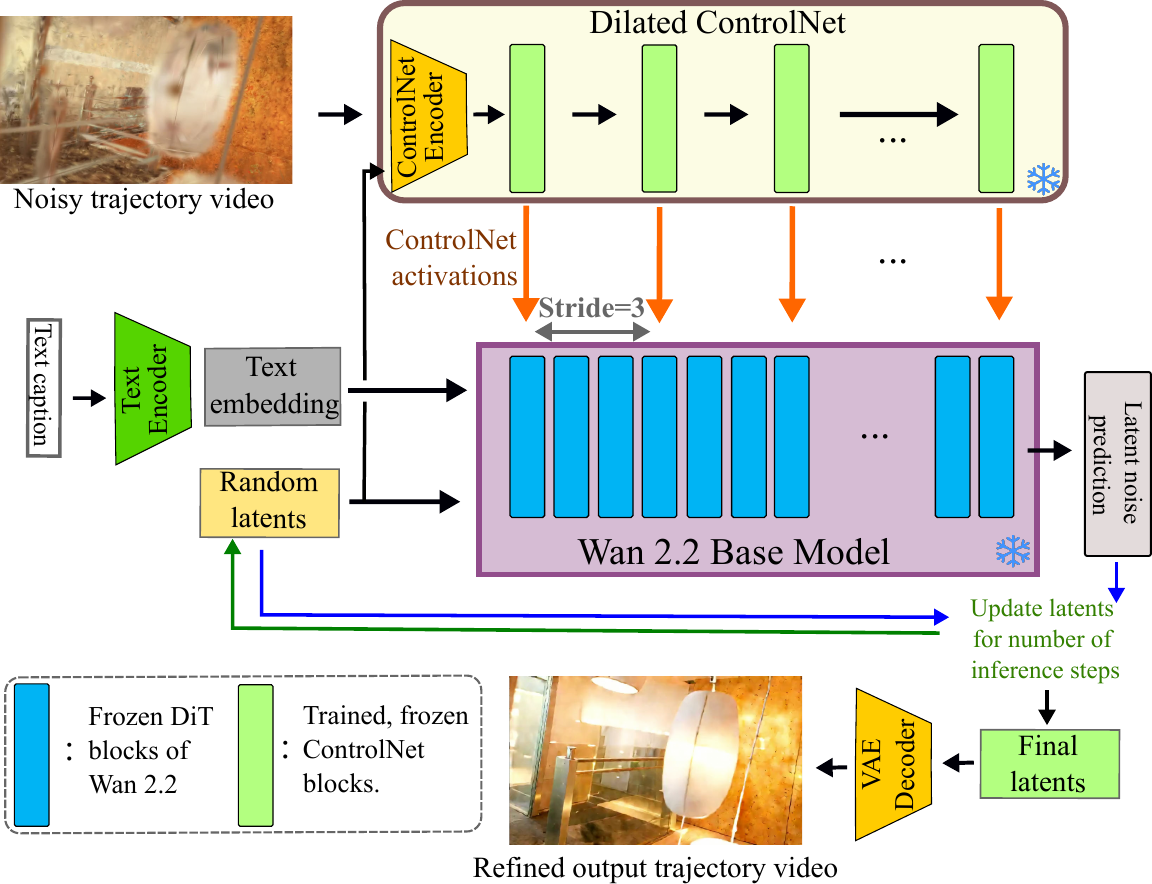}
      \caption{The architecture and inference procedure of our \textbf{Wan + ControlNet (Morph.)} video-to-video diffusion refinement model.}
      \label{fig:supp_arch_test}
\end{figure}

\begin{figure*}[t!]
      \centering
      \includegraphics[scale=0.44]{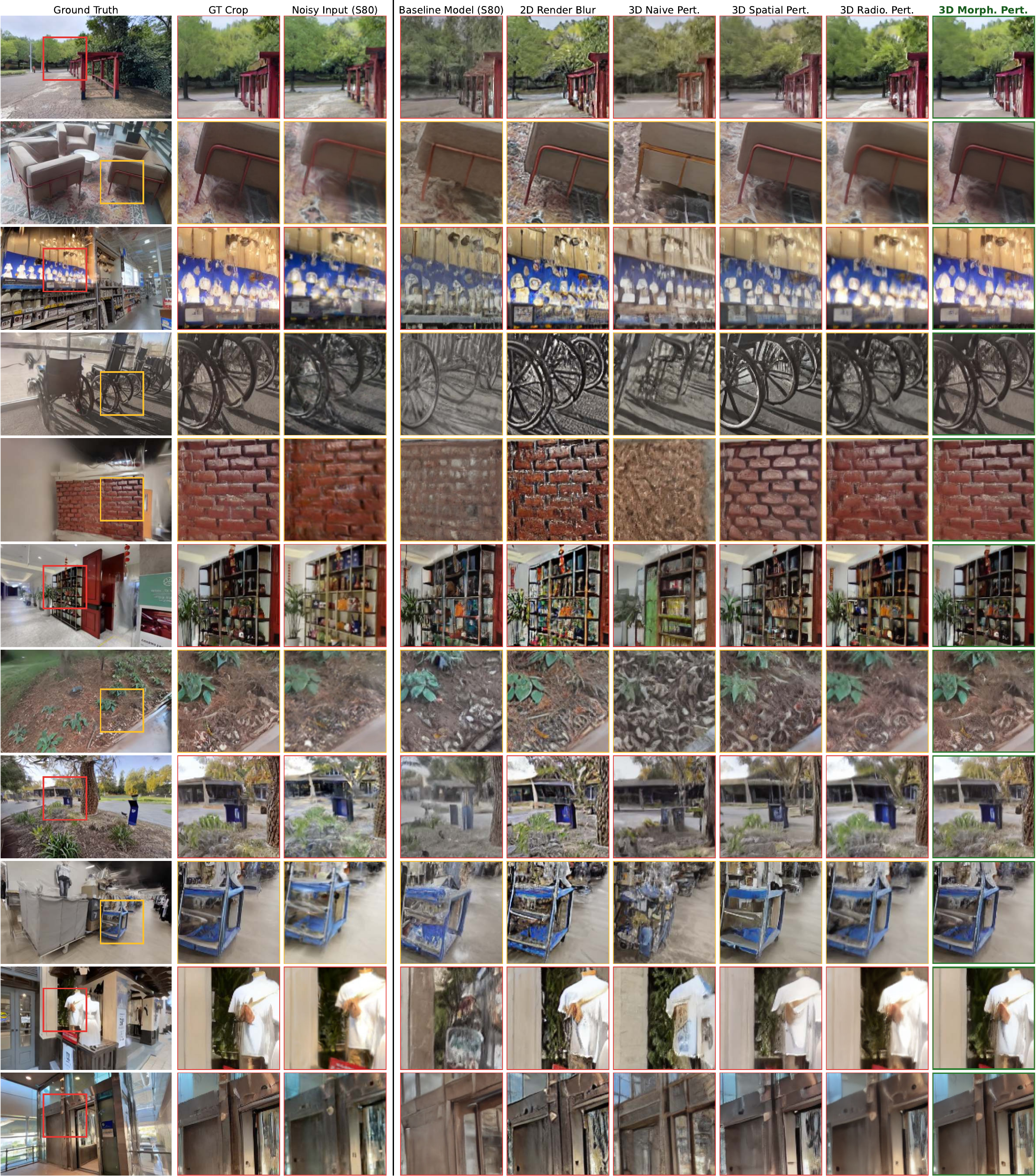}
      %\captionsetup{font=scriptsize}
      \caption{\textbf{Extended qualitative comparison of 3D scenes reconstructed from corrupted video trajectories refined by diffusion models trained on each perturbed dataset.} Evaluated on $\mathcal{S}_{80}$ test set (\textbf{Extreme sparsity}) using 3DGS representations re-optimized from refined output frames.}
      \label{fig:supp_s80}
\end{figure*}

\begin{figure*}[t!]
      \centering
      \includegraphics[scale=0.44]{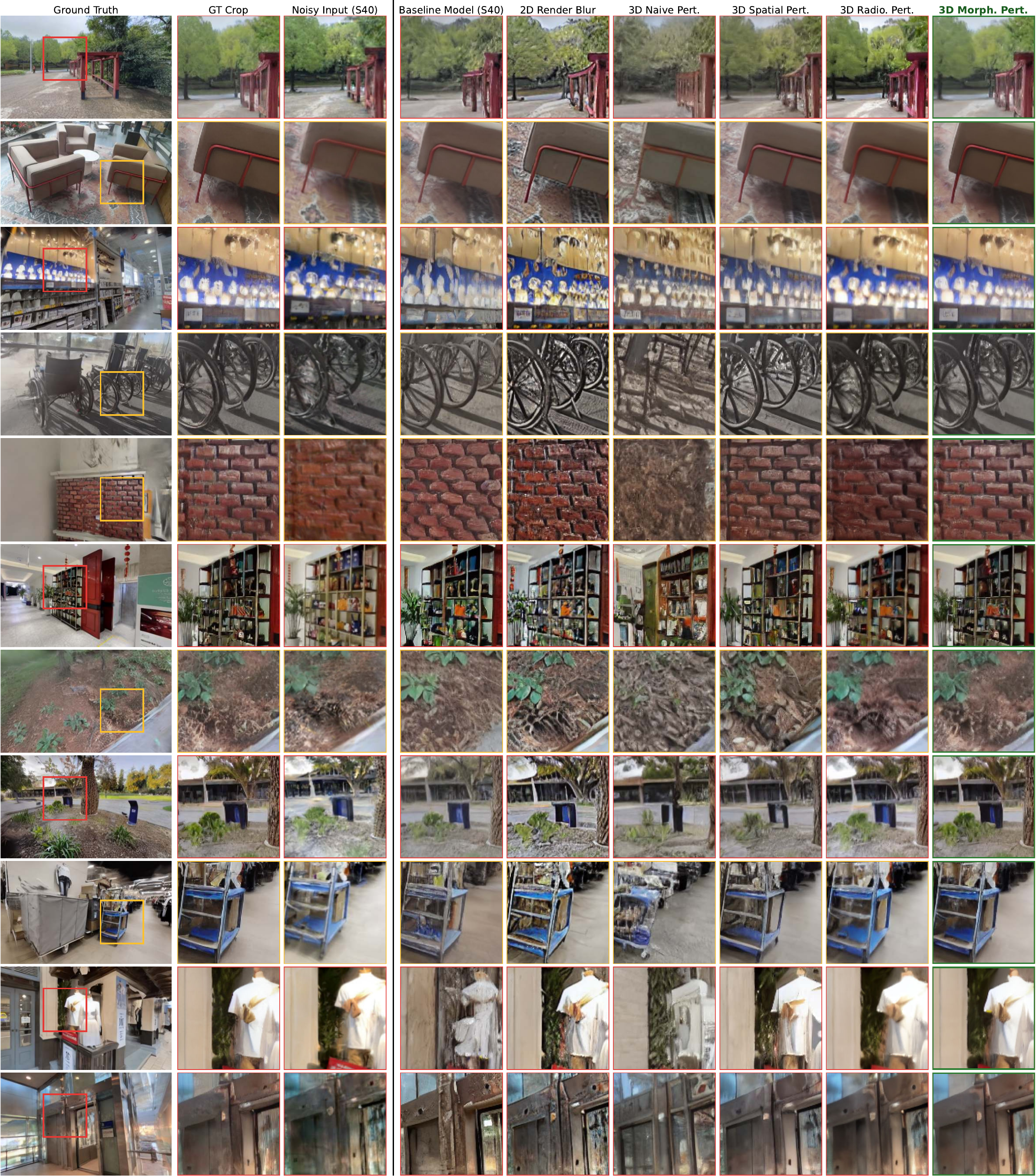}
      %\captionsetup{font=scriptsize}
      \caption{\textbf{Extended qualitative comparison of 3D scenes reconstructed from corrupted video trajectories refined by diffusion models trained on each perturbed dataset.} Evaluated on $\mathcal{S}_{40}$ test set (\textbf{High sparsity}) using 3DGS representations re-optimized from refined output frames.}
      \label{fig:supp_s40}
\end{figure*}

\begin{figure*}[t!]
      \centering
      \includegraphics[scale=0.44]{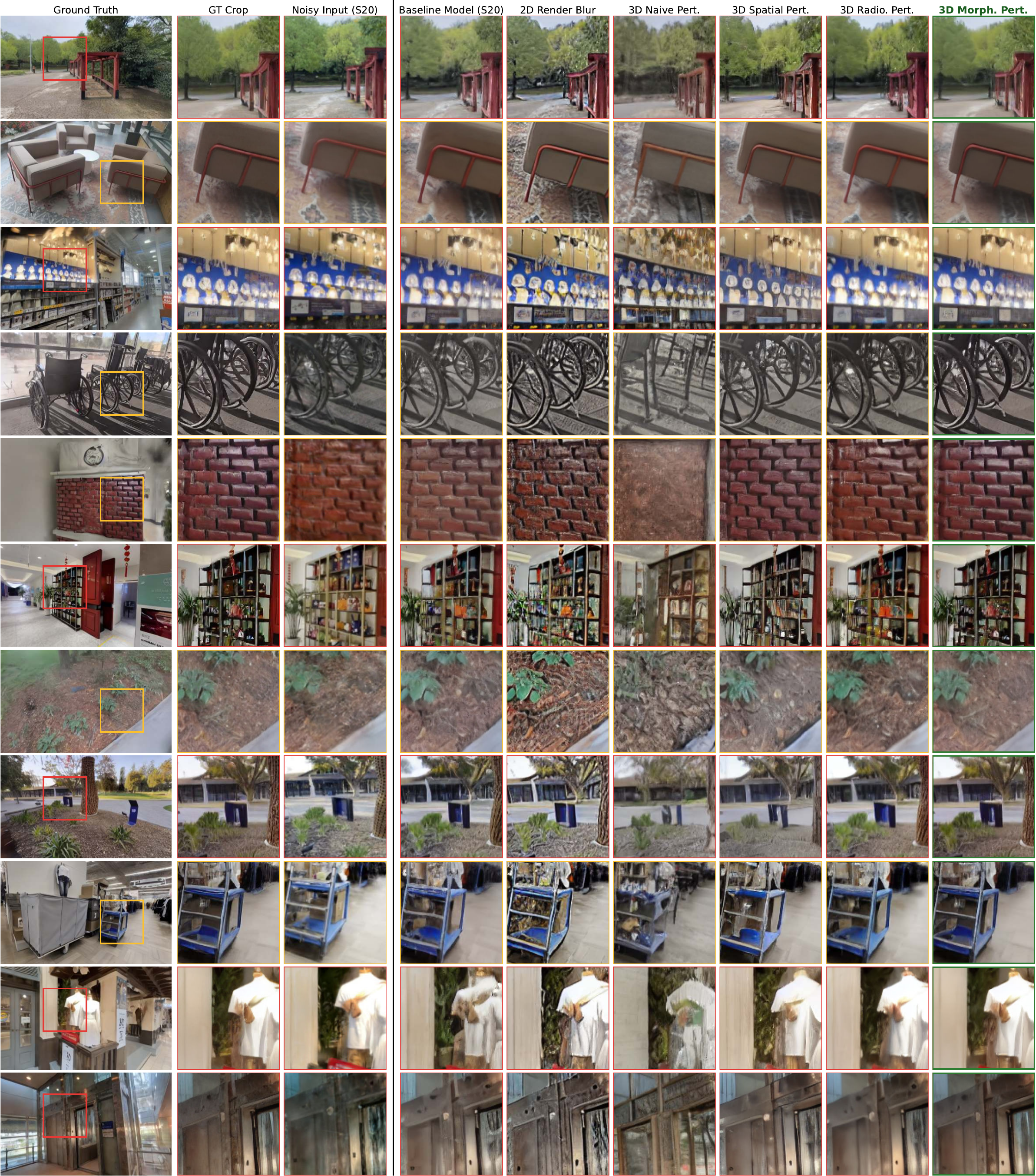}
      %\captionsetup{font=scriptsize}
      \caption{\textbf{Extended qualitative comparison of 3D scenes reconstructed from corrupted video trajectories refined by diffusion models trained on each perturbed dataset.} Evaluated on $\mathcal{S}_{20}$ test set (\textbf{Moderate Sparsity}) using 3DGS representations re-optimized from refined output frames.}
      \label{fig:supp_s20}
\end{figure*}

The inference pipeline is illustrated in Figure~\ref{fig:supp_arch_test}. During inference, the trained ControlNet remains frozen. The input text caption is embedded using Qwen2, while the latent video representation is initialized with random noise $z_T$. The model then iteratively predicts the noise residual at each timestep, which is used by the scheduler to update the latent representation. This denoising process is repeated for 50 inference steps in all experiments.

Finally, the refined latent representation is decoded using Wan2.2's VAE decoder to produce the output trajectory video. To eliminate subtle color saturation shifts between the noisy input trajectory and diffusion output, we perform global color correction in the CIELAB color space [Commission Internationale de l'Éclairage, 1978] by matching the first- and second-order color statistics between the predicted video and the sparse ground-truth images, following standard color transfer approaches [Reinhard \textit{et al.}, 2001].
%~\cite{cielab}
%~\cite{reinhard2001color}

\subsection{Extended Qualitative Results}
\label{sec:wan_additional_results}

\begin{figure*}[t!]
      \centering
      \includegraphics[scale=0.48]{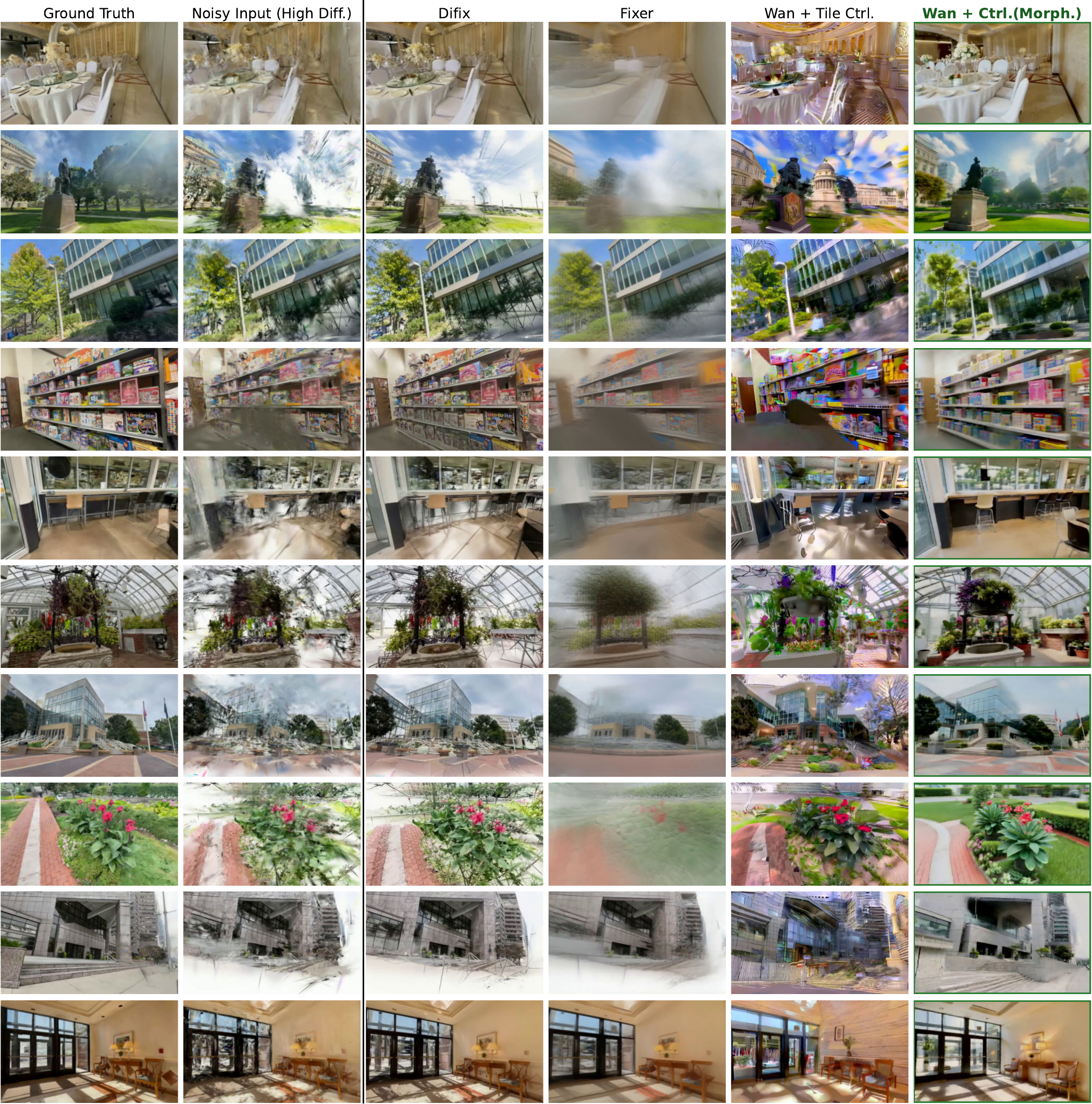}
      %\captionsetup{font=scriptsize}
      \caption{Extended trajectory render refinement qualitative comparison on \textit{High difficulty (6 training views, no overlap w. test trajectory.)} (see Sec. \ref{sec:3dgtr}, \ref{sec:wan_results}) test scenes.}
      \label{fig:supp_6}
\end{figure*}

\begin{figure*}[t!]
      \centering
      \includegraphics[scale=0.48]{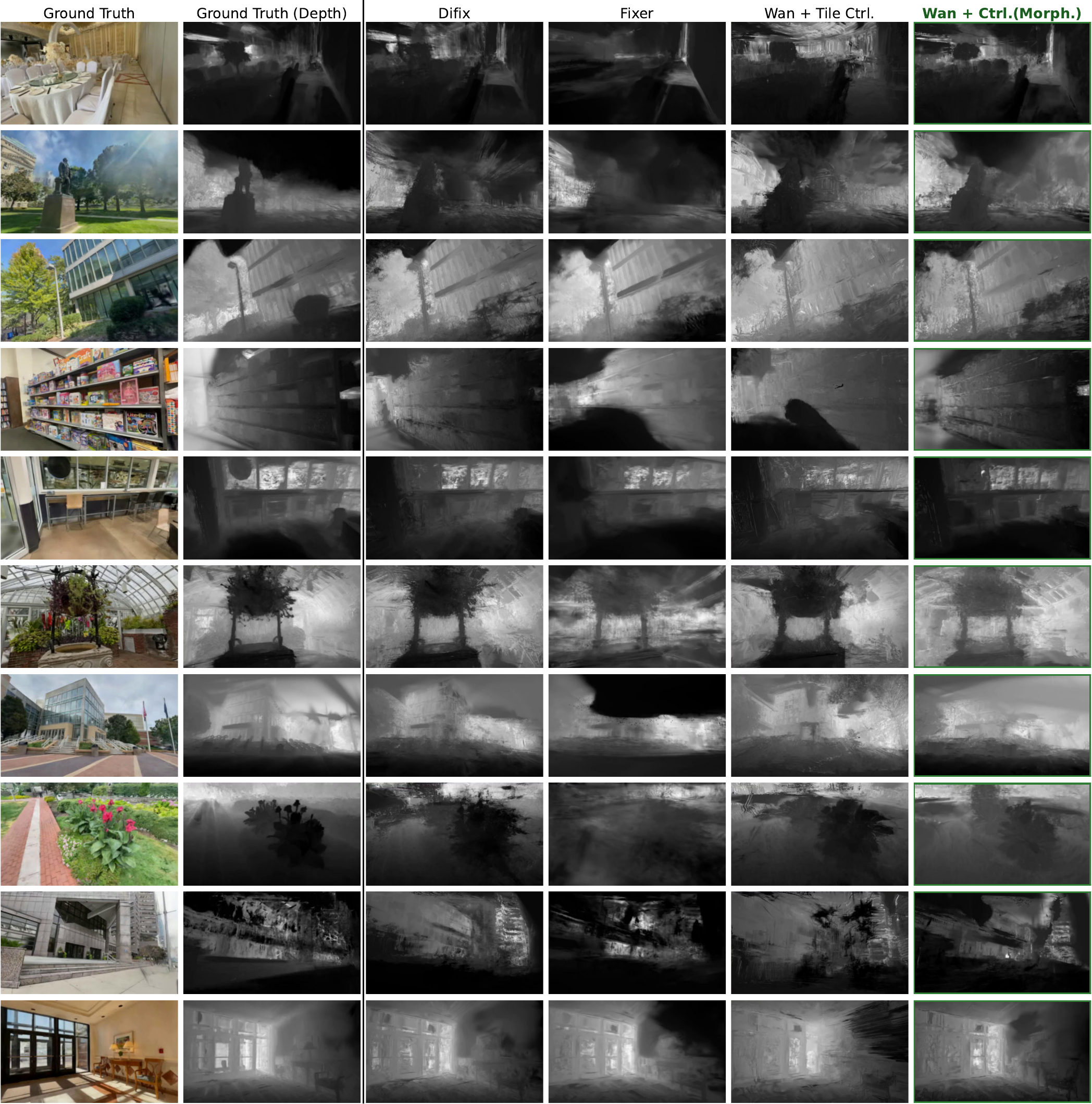}
      %\captionsetup{font=scriptsize}
      \caption{Extended trajectory render refinement qualitative \textbf{depth} comparison on \textit{High difficulty (6 training views, no overlap w. test trajectory.)} (see Sec. \ref{sec:3dgtr}, \ref{sec:wan_results}) test scenes.}
      \label{fig:supp_6_depth}
\end{figure*}

\begin{figure*}[t!]
      \centering
      \includegraphics[scale=0.48]{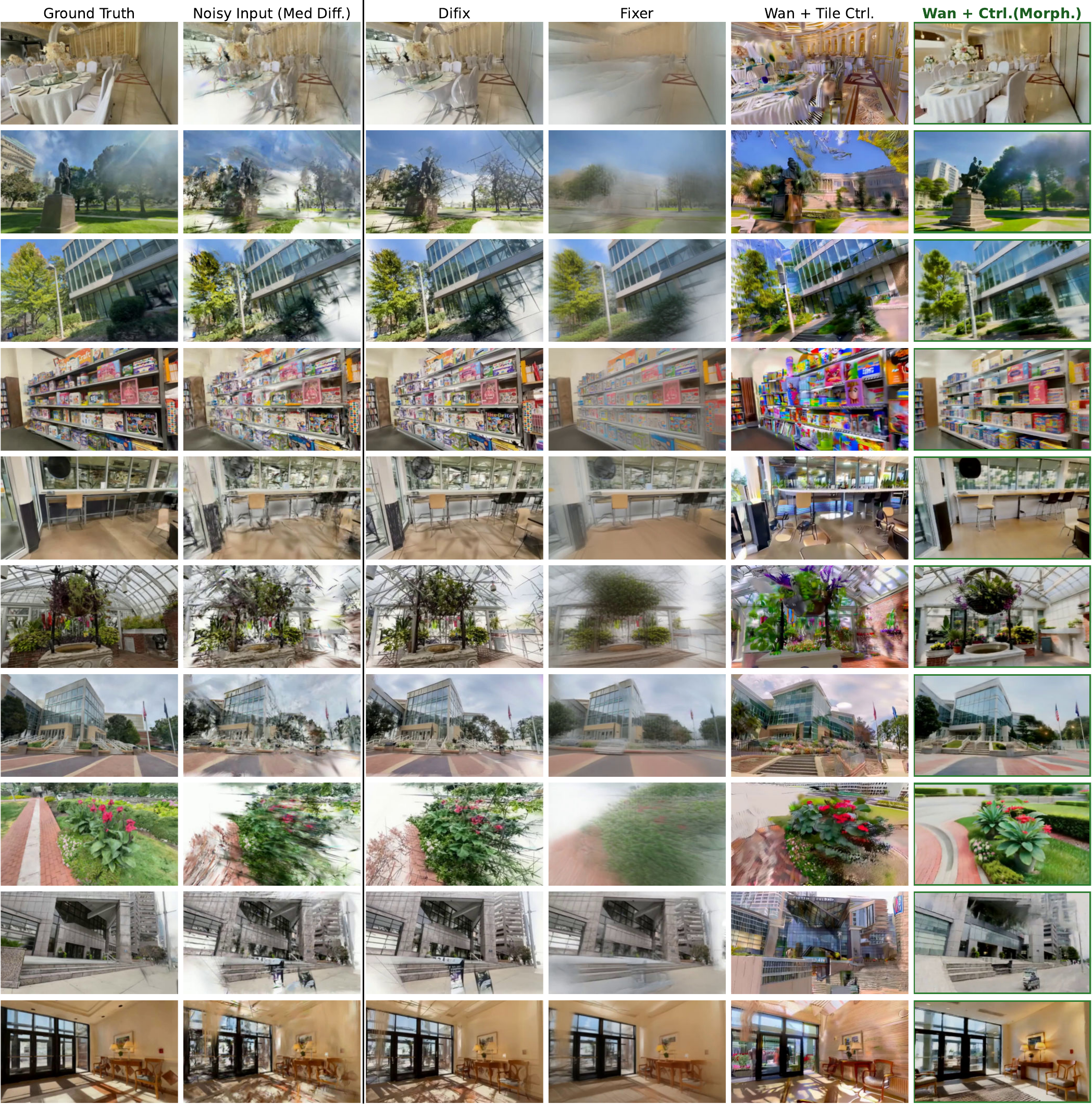}
      %\captionsetup{font=scriptsize}
      \caption{Extended trajectory render refinement qualitative comparison on \textit{Medium difficulty (3 training views, 1 view overlapping w. test trajectory )} (see Sec. \ref{sec:3dgtr}, \ref{sec:wan_results}) test scenes.}
      \label{fig:supp_3}
\end{figure*}

\begin{figure*}[t!]
      \centering
      \includegraphics[scale=0.48]{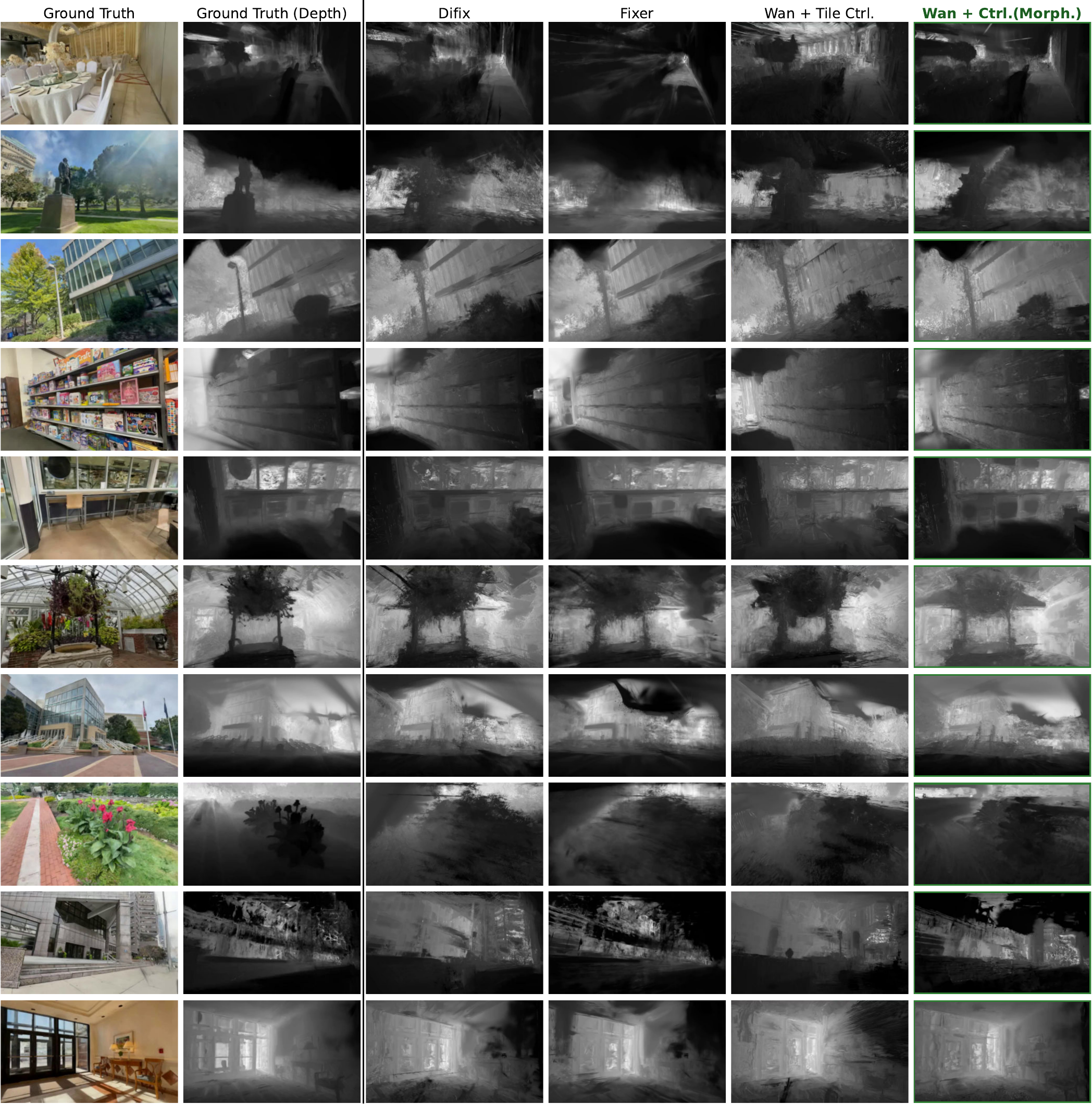}
      %\captionsetup{font=scriptsize}
      \caption{Extended trajectory render refinement qualitative \textbf{depth} comparison on \textit{Medium difficulty (3 training views, 1 view overlapping w. test trajectory )} (see Sec. \ref{sec:3dgtr}, \ref{sec:wan_results}) test scenes.}
      \label{fig:supp_3_depth}
\end{figure*}

\begin{figure*}[t!]
      \centering
      \includegraphics[scale=0.48]{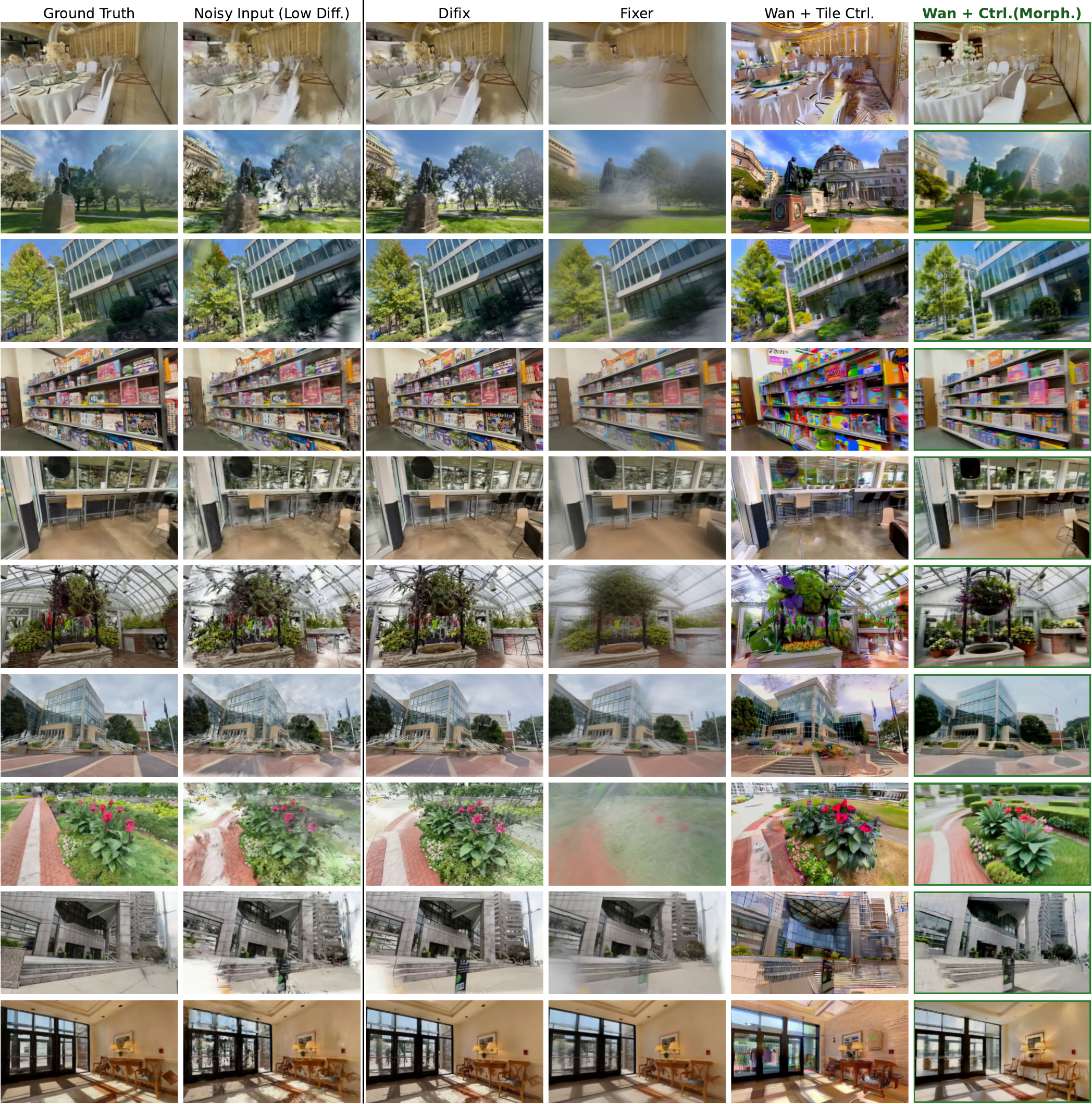}
      %\captionsetup{font=scriptsize}
      \caption{Extended trajectory render refinement qualitative comparison on \textit{Low difficulty (9 training views, 1 view overlapping w. test trajectory )} (see Sec. \ref{sec:3dgtr}, \ref{sec:wan_results}) test scenes.}
      \label{fig:supp_9}
\end{figure*}

\begin{figure*}[t!]
      \centering
      \includegraphics[scale=0.48]{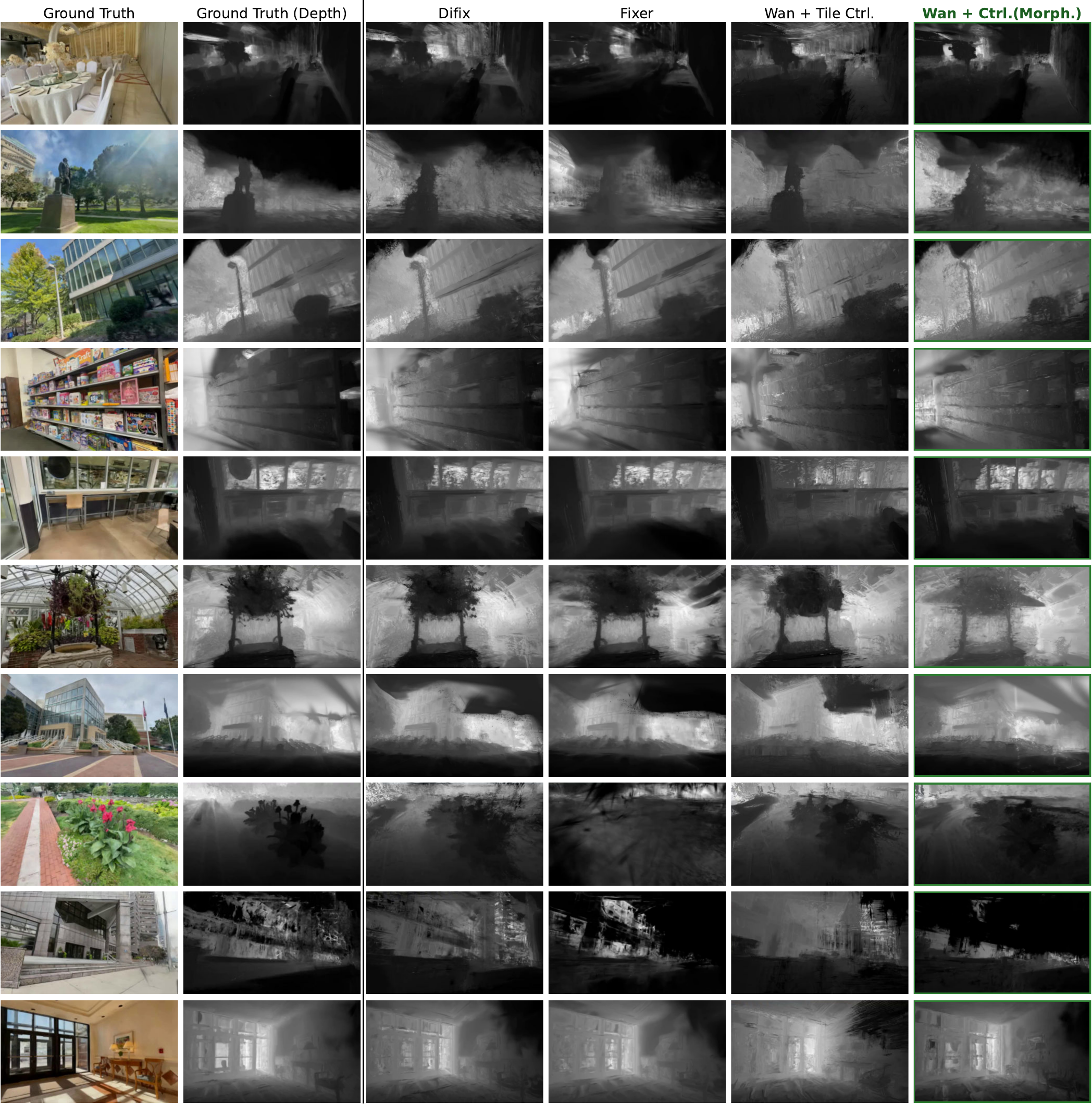}
      %\captionsetup{font=scriptsize}
      \caption{Extended trajectory render refinement qualitative \textbf{depth} comparison on \textit{Low difficulty (9 training views, 1 view overlapping w. test trajectory )} (see Sec. \ref{sec:3dgtr}, \ref{sec:wan_results}) test scenes.}
      \label{fig:supp_9_depth}
\end{figure*}

We extend the trajectory refinement evaluations from Section~\ref{sec:3dgtr} of the main paper with additional qualitative examples on the DL3DV-10K Benchmark~\cite{Ling_2024_CVPR}. Extended visual comparisons across \textit{High} (6-view, no overlap), \textit{Medium} (3-view, 1 overlap), and \textit{Low} (9-view, 1 overlap) difficulty settings are shown in Figures~\ref{fig:supp_6}, \ref{fig:supp_3}, and \ref{fig:supp_9}. We also provide corresponding depth renders in Figures~\ref{fig:supp_6_depth}, \ref{fig:supp_3_depth}, and \ref{fig:supp_9_depth} respectively.

These additional results align with our findings in the main text. Single-frame baselines like \textit{Fixer}~\cite{Wu_2025_CVPR} yield smooth outputs but lose underlying scene structure, whereas \textit{Difix} deteriorates rapidly as multi-view consistency errors increase in harder regimes. Because both lack cross-frame temporal awareness, reconstructing 3DGS representations from their outputs causes severe geometric degradation. Video baselines like \textit{Wan+Tile}~\cite{TheDenk} fail to preserve multi-view geometry, hallucinating details that stray from the ground truth. In contrast, \textbf{Wan + ControlNet (Morph.)} maintains stable visual quality, effectively eliminating rendering artifacts while preserving sharp details and multi-view coherence across all difficulty levels. 

\section{Downstream Robotics Imitation Learning}
\label{sec:drema_supp}

\begin{figure*}[t!]
      \centering
      \includegraphics[scale=0.6]{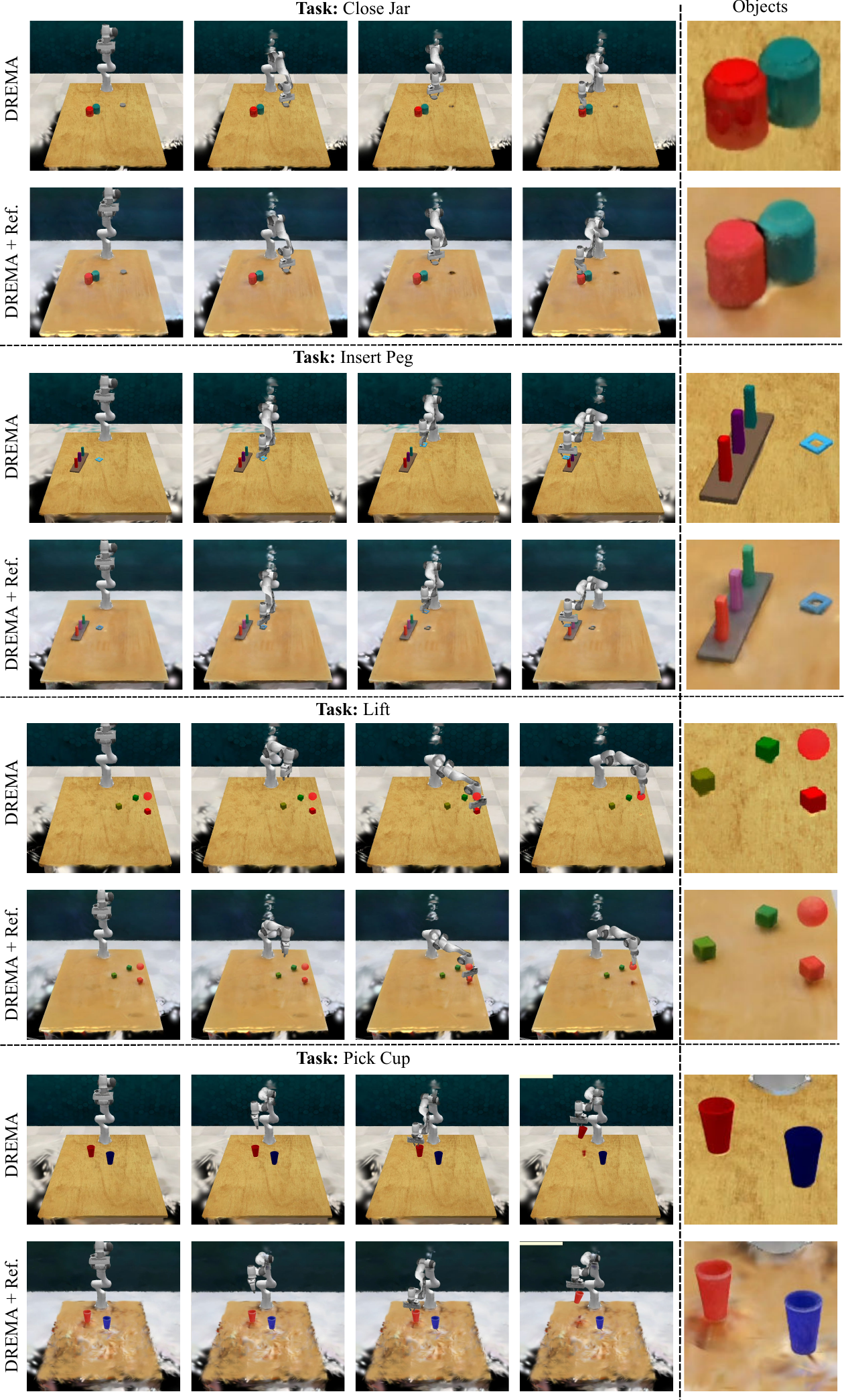}
      %\captionsetup{font=scriptsize}
      \caption{Qualitative comparison of trajectory keyframes for synthetic manipulation task episodes generated by baseline DREMA and \textbf{DREMA + ref.} (\textbf{Wan + ControlNet (Morph.)}) on RLBench tasks, including close-ups of target objects.}
      \label{fig:rl_bench_supp}
\end{figure*}

This section complements Section~\ref{sec:robotics_downstream} of the main paper by providing additional visual comparisons for downstream policy learning on RLBench~\cite{james2019rlbench}. In Section~\ref{sec:robotics_downstream}, we integrate our video refinement model (\textbf{Wan + Ctrl (Morph.)}) into the pipeline of DREMA~\cite{barcellona2025dream}, a framework that reconstructs simulated manipulation tasks as 3D Gaussian Splatting scenes and interfaces with PyBullet~\cite{coumans2016pybullet} physics to recompose scenes into novel task episodes. In Figure~\ref{fig:rl_bench_supp}, we present additional qualitative trajectories across key timesteps for synthetic episodes generated by baseline DREMA and \textbf{DREMA+ref.}, accompanied by zoomed-in views of interaction objects. Consistent with the qualitative findings in Figure~\ref{fig:qualitative_rlbench} of the main paper, these additional visual results demonstrate that our video refinement pipeline introduces photorealistic lighting dynamics, realistic color shifts, and enhanced surface textures to synthetic renders while preserving multi-view and temporal 3D geometric consistency. By augmenting appearance diversity—even in occasional instances where the refinement introduces noisy surfaces, such as the table in the \textit{Pick Cup} task (Figure~\ref{fig:rl_bench_supp})—our approach acts as an effective regularizer, encouraging the downstream model (PerAct~\cite{shridhar2022peract}) to rely on underlying scene geometry during evaluation.

%\caption{\textbf{Extended qualitative comparison of 3D scenes reconstructed from corrupted video trajectories refined by diffusion models trained on each perturbed dataset.} Evaluated on $\mathcal{S}_{80}$ test set (\textbf{High Difficulty}, 6 training views, no overlap w. test trajectory.) using 3DGS representations re-optimized from refined output frames.}




\end{document}